\documentclass[letterpaper, 10 pt, conference]{ieeeconf}  

\IEEEoverridecommandlockouts                              

\input{paper_style.sty}

\newcommand\copyrighttext{%
    \footnotesize 
    \color{gray}
    \textcopyright  
    2025
    \ IEEE. Personal use of this material is permitted.  Permission from IEEE must be obtained for all other uses, in any current or future media, including reprinting/republishing this material for advertising or promotional purposes, creating new collective works, for resale or redistribution to servers or lists, or reuse of any copyrighted component of this work in other works.
}

\title{\LARGE \bf
Clarke Transform and Encoder-Decoder Architecture for\\
Arbitrary Joints Locations in Displacement-Actuated Continuum Robots
}

\author{Reinhard M.~Grassmann and Jessica Burgner-Kahrs
\thanks{We acknowledge the support of the Natural Sciences and Engineering Research Council of Canada (NSERC), [RGPIN-2019-04846].}
\thanks{All authors are with Continuum Robotics Laboratory, Department of Mathematical and Computational Sciences, University of Toronto, Mississauga, ON L5L 1C6, Canada {\tt\small reinhard.grassmann@utoronto.ca}}
}

\begin{document}

\maketitle
\thispagestyle{empty}
\pagestyle{empty}

\begin{abstract}
In this paper, we consider an arbitrary number of joints and their arbitrary joint locations along the center line of a displacement-actuated continuum robot.
To achieve this, we revisit the derivation of the Clarke transform leading to a formulation capable of considering arbitrary joint locations.
The proposed modified Clarke transform opens new opportunities in mechanical design and algorithmic approaches beyond the current limiting dependency on symmetric arranged joint locations.
By presenting an encoder-decoder architecture based on the Clarke transform, joint values between different robot designs can be transformed enabling the use of an analogous robot design and direct knowledge transfer.
To demonstrate its versatility, applications of control and trajectory generation in simulation are presented, which can be easily integrated into an existing framework designed, for instance, for three symmetric arranged joints.
\end{abstract}

\begin{tikzpicture}[remember picture,overlay]
        \node[anchor=south,yshift=10pt] at (current page.south) {\parbox{\dimexpr0.75\textwidth-\fboxsep-\fboxrule\relax}{\copyrighttext}};
\end{tikzpicture}

\section{Introduction}

Real-world tasks in medical and industrial applications do not impose symmetric manipulator designs for soft and continuum robots.
However, almost all displacement-actuated continuum robots rely on a very narrow design space, \textit{i.e.}, three or four symmetric arranged joints.
Overcoming that design restriction has obvious benefits.
In particular, increasing the joint number $n$ significantly, \textit{i.e.}, $n\!\gg\!3$, and considering asymmetric joint arrangements: (i) enhances manipulability along the directions that coincide with joint locations and center-line; (ii) improves force absorption and delivery due to the distributed nature of the actuation forces; and (iii) increases safety through actuation redundancy. 
Desirable in medical applications \cite{Burgner-KahrsRuckerChoset_TRO_2015, DupontRucker_et_al_JPROC_2022} and industrial settings \cite{DongKell_et_al_JMP_2019, RussoAxinte_et_al_AIS_2023}, the utilization of such mechanical design may lead to increased load capacity, better shape conformation, enhanced stability, and variable stiffness.
Therefore, exploring and providing a computational inexpensive general approach is a pioneering step towards more capable displacement-actuated continuum robots that include a large set of continuum and soft robots.

To this day, displacement-actuated continuum robots with $n$ joints are under-explored.
Some attempts, \textit{e.g.}, \cite{LuWang_et_al_AR_2020, AllenAlbert_et_al_RoboSoft_2020} for $n$ joints provide the solution for the robot-independent mapping, and the work by Dalvand \textit{et al.}
\cite{DalvandNahavandiHowe_Access_2022} considers $n$ joints in their forward kinematics framework.
Their framework includes an exhaustive list of $2^n$ combinations of passive and active tendons as joints and several branches, \textit{e.g.}, if-else-statements, as well as function calls to numerically solve an underlying beam model.
However, a computed result might not be consistent, where a consistency between \SI{96.1}{\%} and \SI{99.2}{\%} is reported.
While this approach \cite{DalvandNahavandiHowe_Access_2022} can be considered to account for arbitrary joint locations, it relies heavily on computing and heuristics.

\begin{figure}
    \centering
    \vspace{0.75em}
    \begin{overpic}[width=\columnwidth]{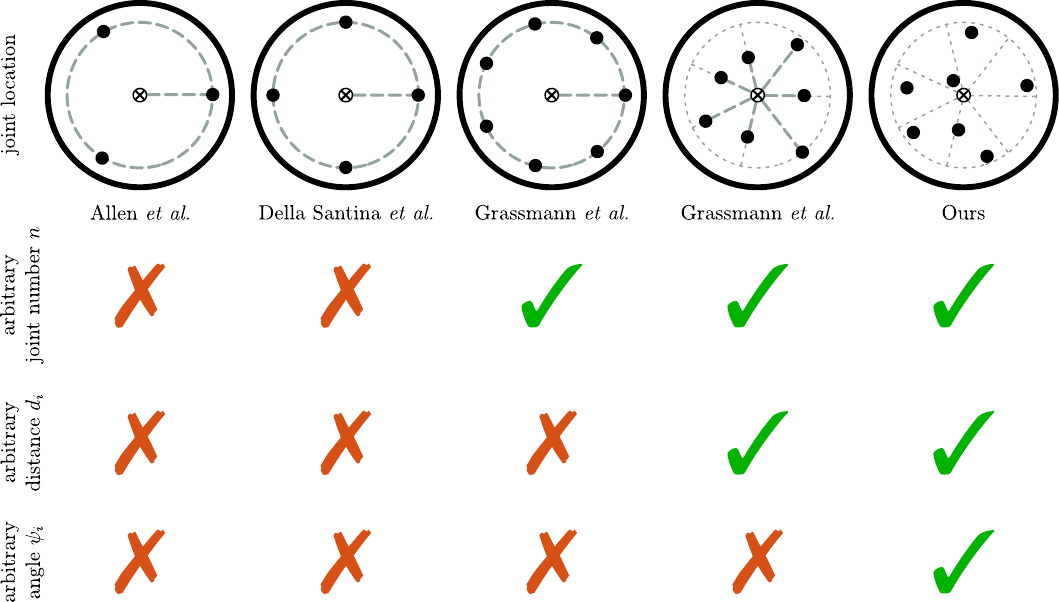}
        \put(18.5,36.125){\tiny{\cite{AllenAlbert_et_al_RoboSoft_2020}}}
        \put(41.5,36.125){\tiny{\cite{DellaSantinaBicchiRus_RAL_2020}}}
        \put(60,36.125){\tiny{\cite{GrassmannSenykBurgner-Kahrs_arXiv_2024}}}
        \put(79.4,36.125){\tiny{\cite{GrassmannBurgner-Kahrs_ICRA_EA_2024}}}
    \end{overpic}
    \caption{
        Joint location and improved joint representation.
        The kinematics of a displacement-actuated continuum robot with fixed segment length is mainly influenced by the number of joints $n$ and their location in the cross-section in terms of polar coordinates.
        For the $i\textsuperscript{th}$ joint, the polar coordinates is described by the distance $d_i$ to the center-line and the angle $\psi_i$.
        Joint representations, \textit{e.g.}, \cite{AllenAlbert_et_al_RoboSoft_2020, DellaSantinaBicchiRus_RAL_2020, GrassmannSenykBurgner-Kahrs_arXiv_2024, GrassmannBurgner-Kahrs_ICRA_EA_2024}, have been proposed to consider various arrangement.
        Our approach generalizes and covers all cases.
        }
    \label{fig:catchy}
    \vspace*{-1.5em}
\end{figure}
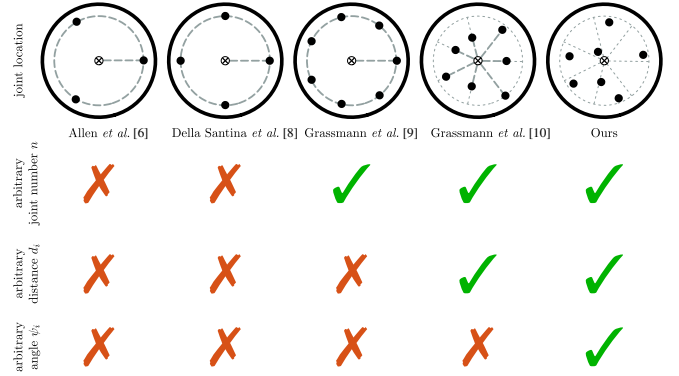

In our recent work \cite{GrassmannSenykBurgner-Kahrs_arXiv_2024}, they show that the Clarke transform using a generalized Clarke transformation matrix can be used to map $n$ joint values onto a two-dimensional manifold.
Kinematics based on the Clarke transform are branchless, closed-form, and singularity-free.
However, this approach \cite{GrassmannSenykBurgner-Kahrs_arXiv_2024} only applies to symmetric arranged joint location, see Fig.~\ref{fig:catchy}.
The Clarke transform can also account for non-constant distances \cite{GrassmannBurgner-Kahrs_ICRA_EA_2024}.
However, to the best of our knowledge, no approach can consider arbitrary joint location while only considering the kinematic design parameters of the target displacement-actuated continuum robot.
The kinematic design parameters entail angular offset $\psi_i$, distance $d_i$ to center-line in the cross-section, and length $l$ of a displacement-actuated continuum robot as illustrated in Fig.~\ref{fig:design_parameters}.

Due to the Clarke transform's relationship to the Clarke transformation matrix, looking into generalized Clarke transformation matrices for $n$ phases in the literature for electrical motors is worthwhile.
Furthermore, based on the analogy to Kirchhoff's current law \cite{GrassmannSenykBurgner-Kahrs_arXiv_2024}, the actuation constraint, \textit{i.e.}, $\sum_{i=1}^n \rho_i = 0$, where $\rho_i$ are the displacement values, inherent to a displacement-actuated continuum robot with symmetric arrangement can be considered as balanced system.
Consequently, a displacement-actuated continuum robot with asymmetric arrangement can be considered an unbalanced system as the constraint is not necessarily zero akin to an unbalanced electrical system, where the sum of all electrical current in each phase is non-zero.

Janaszek \cite{Janaszek_PIE_2016} derives a similar matrix to ours \cite{GrassmannBurgner-Kahrs_ICRA_EA_2024, GrassmannSenykBurgner-Kahrs_arXiv_2024}, which differs only by a scaling factor.
In work by Willems \cite{Willems_TOE_1969} and by Rockhill \& Lipo \cite{RockhillLipo_IEMDC_2015}, the squared matrix is only useful for balanced systems.
For symmetric phase arrangements, all variants should produce $\left(n - 2\right)$ zeros resulting in unnecessary computation and large state representation.
Another squared matrix is derived by Willems \cite{Willems_TOE_1969}, which can consider unbalanced systems.
However, for all representations derived by Willems \cite{Willems_TOE_1969}, it is not guaranteed that its inverse produces always $\left(n - 2\right)$ zeros.
Therefore, the dimensionality cannot be reduced to two variables.
To summarize, none of the generalized Clarke transformation matrices can be used directly.

In the present work, we propose a modified generalized Clarke transformation matrix to construct a Clarke transform that applies to arbitrary joint locations for displacement-actuated continuum robots.
Furthermore, we propose an encoder-decoder architecture based on the Clarke transform to facilitate the use of those continuum robots.
In particular, the contribution of this paper includes:
\begin{itemize}
    \item Deriving a generalized Clarke transformation matrix for unbalanced systems
    \item Modifying the Clarke transform to consider arbitrary joint locations
    \item Proposing an encoder-decoder architecture to transform joint values between different robot designs
    \item Adapting a $\mathcal{C}^4$-smooth trajectory generator
    \item Generating feasible joint values 
\end{itemize}
Due to the use of Clarke coordinates, this approach automatically contributes to providing robot-dependent mapping. 
Furthermore, all derived approaches are linear, compact, and closed-form.
\section{Extending Clarke Transform}

In this section, we briefly revisit the Clarke transform and state the relationship to arc parameters.
Afterward, an alternative derivation is stated that does not rely on the connection to a Clarke transformation matrix.
Based on the alternative derivation, we derive a Clarke transform and its inverse to consider asymmetric arranged joint locations.

\subsection{Clarke Transform for Displacement-Actuated Joint}

Using the representation proposed in \cite{GrassmannSenykBurgner-Kahrs_arXiv_2024} for $i\textsuperscript{th}$ entry of displacement-actuated joints $\rhovec$ given by $\rho_i$, the index notation for $\rhovec$ is
\begin{align}
    \rhovec = \left( \rho_\text{Re}\cos{\psi_i} + \rho_\text{Im}\sin{\psi_i} \right)_i \subset \mathbb{R}^n,
    \label{eq:rho}
\end{align}
whereas Clarke coordinates, \textit{i.e.}, $\rhoreal$ and $\rhoim$, being the free parameter of \eqref{eq:rho}, can be combined into
\begin{align}
    \rhoclarke = \left[ \rhoreal, \rhoim \right]\transpose \in \mathbb{R}^2.
    \label{eq:rho_clarke}
\end{align}
Note that \eqref{eq:rho} is an element of a subset of $\mathbb{R}^n$ since all joints are interdependent and constrained.
Therefore, not all elements of $\mathbb{R}^n$ can represent \eqref{eq:rho}.

Both representations can be transformed into each other using
\begin{align}
    \rhoclarke &= \MP\rhovec
    \quad\text{and}\quad
    \label{eq:forward}
    \\
    \rhovec &= \MPinv\rhoclarke,
    \label{eq:inverse}
\end{align}
where $\MPinv$ is the right-inverse of $\MP$ and $\MP$ is given by the generalized Clarke transformation matrix.
For a symmetric arranged joint location, we kindly refer to \cite{GrassmannSenykBurgner-Kahrs_arXiv_2024} for a derivation and specific realization of both matrices.

\subsection{Relation to Arc Parameters}

\begin{figure}
    \centering
    \vspace*{0.75em}
    \includegraphics[width=0.8\columnwidth]{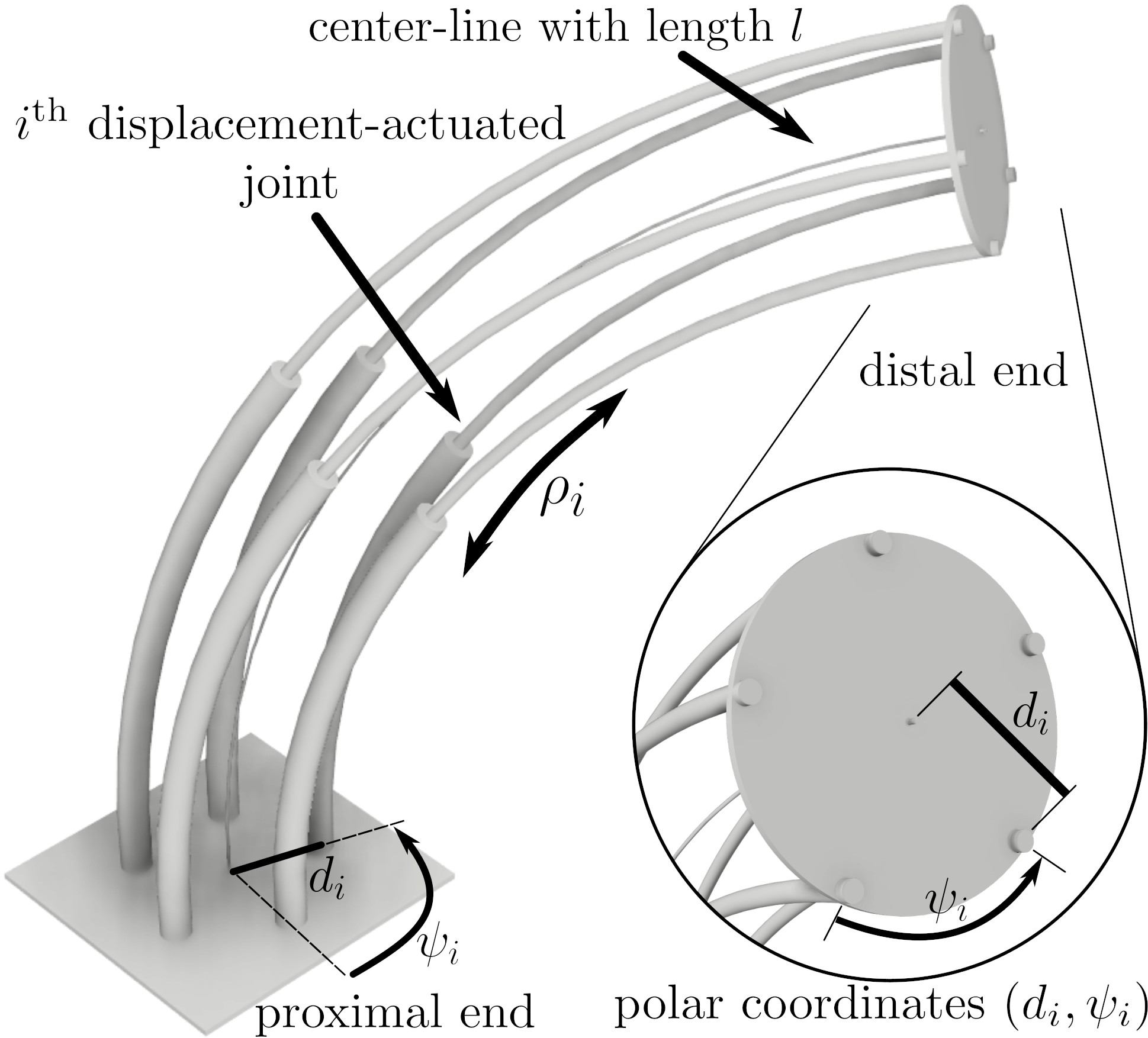}
    \vspace*{-0.5em}
    \caption{
        Kinematic design parameters of a displacement-actuated continuum robot.
        (Image credit: \textit{Grassmann et al.} \cite{GrassmannSenykBurgner-Kahrs_arXiv_2024})
        }
    \label{fig:design_parameters}
    \vspace*{-1.5em}
\end{figure}

We pointed out in \cite{GrassmannBurgner-Kahrs_ICRA_EA_2024} that it is possible to normalize the Clarke coordinates with respect to kinematic design parameters, \textit{i.e.}, distance $d_i$ and segment length $l$, which leads to the curvature-curvature representation if constant curvature is assumed, \textit{i.e.},
\begin{align}
    \begin{bmatrix}
        \kappa\cos\left(\theta\right) \\ \kappa\sin\left(\theta\right)
    \end{bmatrix}
    = 
    \underbrace{
    1 /\, l
    }_{\substack{\text{removes}\ l}}
    \overbrace{
    \boldsymbol{M}_\mathcal{P}
    }^{\substack{\text{removes}\ \psi_i}}
    \underbrace{
    \diag\left(1/d_i\right)
    }_{\substack{\text{removes}\ d_i}}
    \boldsymbol{\rho}
    .
    \label{eq:robot_dependent_mapping_forward}
\end{align}
For its inverse, the parameters $l$, $d_i$, and $\psi_i$ are added, \textit{i.e.},
\begin{align}
    \boldsymbol{\rho}
    = 
    \underbrace{
    l
    }_{\substack{\text{adds}\ l}}
    \overbrace{
    \diag\left(d_i\right)
    }^{\substack{\text{adds}\ d_i}}
    \underbrace{
    \boldsymbol{M}_\mathcal{P}^{-1}
    }_{\substack{\text{adds}\ \psi_i}}
    \begin{bmatrix}
        \kappa\cos\left(\theta\right) \\ \kappa\sin\left(\theta\right)
    \end{bmatrix}
    .
    \label{eq:robot_dependent_mapping_inverse}
\end{align}
For both formulations, the assumption $d_i = d$ has been removed. 
The kinematic design parameters fully describe the displacement-actuated continuum robot in the kinematic sense, see Fig.~\ref{fig:design_parameters}.

\subsection{Alternative Derivation of $\MP$}

In our prior work \cite{GrassmannSenykBurgner-Kahrs_arXiv_2024}, we generalize the Clarke transformation matrix motivated by an analogy between a tendon-driven continuum robot and the control of brushless motors.
In fact, the more important part is joint representation \eqref{eq:rho}.
To show this, we present an alternative approach to derive $\MP$.

First, the Clarke coordinates in \eqref{eq:rho} are factored out, \textit{i.e.},
\begin{align}
    \begin{bmatrix} 
        \rho_1 \\ \rho_2 \\ \vdots \\ \rho_n 
    \end{bmatrix}
    =
    \underbrace{
	\begin{bmatrix}
		\cos\left(\psi_1\right) & \sin\left(\psi_1\right) \\
        \cos\left(\psi_2\right) & \sin\left(\psi_2\right) \\
        \vdots & \vdots\\
        \cos\left(\psi_n\right) & \sin\left(\psi_n\right)
	\end{bmatrix}
    }_{\substack{\MPinv}}
    \begin{bmatrix} 
        \rhoreal \\ \rhoim
    \end{bmatrix}
    ,
	\label{eq:MP_inverse}
\end{align}
which resembles \eqref{eq:inverse}.
As a result, a generalized inverse Clarke transformation matrix denoted by $\MPinv$ can be identified.
For the sake of clarification, the notation used for $\MPinv$ does not indicate a matrix inverse of $\MP$.

Second, constructing the Moore-Penrose pseudoinverse for solving undetermined linear systems leads to \eqref{eq:forward}, where
\begin{align}
    \MP =& \left(\left(\MPinv\right)\transpose\MPinv\right)^{-1}\left(\MPinv\right)\transpose.
    \label{eq:MP_inverse_pseudo}
\end{align}
Note that \eqref{eq:MP_inverse_pseudo} is not an approximation and leads to an exact solution, because an underdetermined linear system is solved.

Finally, we assume that the angle $\psi_i$ represented a symmetric arrangement of the joint locations, \textit{i.e.}, $\psi_i = 2\pi (i - 1) / n$.
For this specific case, we can write
\begin{align}
    \rhoclarke =& \left(\left(\MPinv\right)\transpose\MPinv\right)^{-1}\left(\MPinv\right)\transpose\rhovec
    \nonumber\\
    =& \left(
    \begin{bmatrix}
        n/2 & 0\\
        0 & n/2\\
    \end{bmatrix}
    \right)^{-1}
    \left(\MPinv\right)\transpose\rhovec
    =
    \,
    \dfrac{2}{n}\left(\MPinv\right)\transpose\rhovec
    ,
    \nonumber
\end{align}
where the trigonometric identity, \textit{i.e.}, $\sum_{i=1}^{n} \sin^2\left(\psi_i\right) = {n}/{2}$,
$\sum_{i=1}^{n} \cos^2\left(\psi_i\right) = {n}/{2}$, and $\sum_{i=1}^{n} \sin\left(\psi_i\right)\cos\left(\psi_i\right) = 0$, derived and stated in \cite{GrassmannBurgner-Kahrs_ICRA_EA_2024, GrassmannSenykBurgner-Kahrs_arXiv_2024} are used to rewrite $\left(\MPinv\right)\transpose\MPinv$.
Per definition, $\MP$ is set to ${2}/{n}\left(\MPinv\right)\transpose$, which aligns with the property stated in \cite{GrassmannSenykBurgner-Kahrs_arXiv_2024}.

\subsection{Asymmetric arranged joint locations}

\begin{figure}[thb]
    \centering
    \includegraphics[width=\columnwidth]{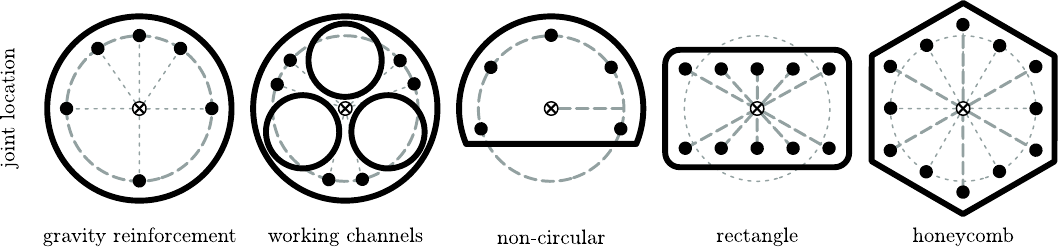}
    \caption{
        Possible designs of arbitrary asymmetric joint location.
        }
    \label{fig:non-symmetric_joint-location}
\end{figure}

In order to consider asymmetric designs such as illustrated in Fig.~\ref{fig:non-symmetric_joint-location}, the Clarke transform \cite{GrassmannSenykBurgner-Kahrs_arXiv_2024} can be modified. 
The pseudoinverse \eqref{eq:MP_inverse_pseudo} and inverse robot-dependent mapping \eqref{eq:robot_dependent_mapping_inverse} point towards the possibility to relax assumption $\psi_i = 2\pi (i - 1) / n$ and $d_i = d$, respectively.

A modification leads to
\begin{align}
    \MPinvalt = \dfrac{1}{f\!\left(d_i\right)} \diag\left(d_i\right) \MPinv ,
    \nonumber
\end{align}
where $\MPalt$ can be found using the Moore-Penrose pseudoinverse considering the whole right-hand side of expression akin to \eqref{eq:MP_inverse_pseudo}.
The diagonal matrix $\diag\left(d_i\right) \in \mathbb{R}^n$ is used to rescale $\rhoclarke$, \textit{cf.} \eqref{eq:robot_dependent_mapping_forward} and \eqref{eq:robot_dependent_mapping_inverse}.
To account for the normalization and change of unit due to $\diag\left(d_i\right)$, the function $f\!\left(d_i\right)$ is introduced and outputs a scalar for a given list or vector of all $d_i$.
The unit of the scalar is a unit of length.
However, many different choices can be justified that are dependent on a specific application.
To overcome the choice of a suitable function $f\!\left(d_i\right)$, we propose an encoder-decoder architecture. 
\section{Encoder-Decoder Architecture}

\begin{figure}
    \centering
    \vspace*{0.75em}
    \includegraphics[width=\columnwidth]{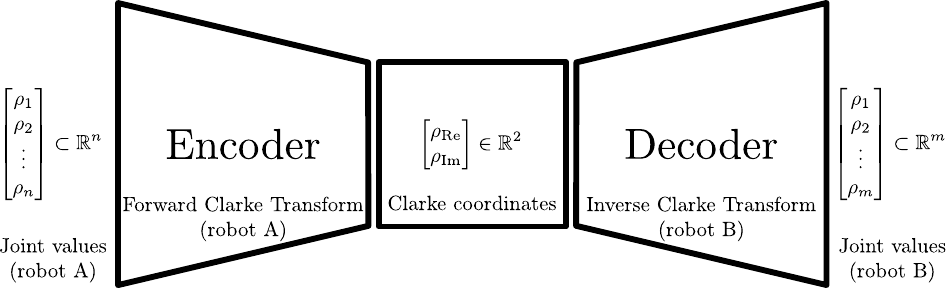}
    \vspace*{-1.5em}
    \caption{
        Encoder-decoder architecture.
        Joint values of one robot type (robot A) with $n$-dimensional joint space can be transformed into joint values of a different robot type (robot B) with $m$-dimensional joint space.
        The latent space representation is encoded as Clarke coordinates.
        It is worth noticing that the compression is a lossless compression that allows joint values to be uniquely reconstructed from the Clarke coordinates.
        }
    \label{fig:encoder-decoder}
\end{figure}

The Clarke transform is applicable to a wide variety of different robot morphologies.
Furthermore, the Clarke coordinates are generalized improved state representations \cite{GrassmannSenykBurgner-Kahrs_arXiv_2024}.
With that in mind, the joint values $\rhovec_{\left(\text{robot A}\right)}$ of \textit{robot A} with specific number of joints can be mapped to the same Clarke coordinates $\rhoclarke$, that correspond to the joint values $\rhovec_{\left(\text{robot B}\right)}$ of \textit{robot B} with a different number of joints.
This can be expressed by
\begin{align}
    \rhoclarke = {\MP}_{\left(\text{robot A}\right)}\rhovec_{\left(\text{robot A}\right)} = {\MP}_{\left(\text{robot B}\right)}\rhovec_{\left(\text{robot B}\right)}
    .
    \nonumber
\end{align}
Rearranging the above equation leads to an \textit{encoder-decoder architecture} illustrated in Fig.~\ref{fig:encoder-decoder} and is given by
\begin{align}
    \rhovec_{\left(\text{robot B}\right)} = 
    \underbrace{
    {\MPinv}_{\left(\text{robot B}\right)}
    }_{\substack{\text{decoder}}}
    \overbrace{
    {\MP}_{\left(\text{robot A}\right)}
    }^{\substack{\text{encoder}}}
    \rhovec_{\left(\text{robot A}\right)},
    \label{eq:encoder-decoder}
\end{align}
which assumed symmetric arranged joint locations.
Note that ${\MPinv}$ is the right-inverse of ${\MP}$, \textit{i.e.}, ${\MPinv}{\MP} \not= \boldsymbol{I}$.
Therefore, \eqref{eq:encoder-decoder} is derived by using \eqref{eq:forward} and \eqref{eq:inverse} directly.

To relax the assumption, \eqref{eq:robot_dependent_mapping_forward} and \eqref{eq:robot_dependent_mapping_inverse} can be used to remove and add the kinematic design parameters, respectively.
In this case, \eqref{eq:robot_dependent_mapping_forward} is the encoder, whereas \eqref{eq:robot_dependent_mapping_inverse} is the decoder.
This results in
\begin{align}
    \begin{split}
        \boldsymbol{\rho}_{\left(\text{robot B}\right)}
        =&
        \overbrace{
        l_{\left(\text{B}\right)}
        \diag\left(d_{i, \left(\text{B}\right)}\right)
        {\MPinv}_{\left(\text{B}\right)}
        }^{\substack{\text{adds kinematic design parameters of robot B}}}
        \\
        &\cdot
        \hspace{-1.5em}
        \underbrace{
        \dfrac{1}{l_{\left(\text{A}\right)}}
        {\MP}_{\left(\text{A}\right)}
        \diag\left(\dfrac{1}{d_{i, \left(\text{A}\right)}}\right)
        }_{\substack{\text{removes kinematics design parameters of robot A}}}
        \hspace{-1.5em}
        \boldsymbol{\rho}_{\left(\text{robot A}\right)}
        ,
    \end{split}
    \label{eq:encoder-decoder_general}
\end{align}
where, for brevity, the subscript \textit{A} and \textit{B} is short for \textit{robot A} and \textit{robot B}, respectively.
The generalized Clarke transformation matrix $\MP$ in \eqref{eq:encoder-decoder_general} is defined by \eqref{eq:MP_inverse_pseudo}.

\section{Demonstration in Simulation}

\begin{figure*}[thb]
    \centering
    \vspace*{0.75em}
    \includegraphics[width=\textwidth]{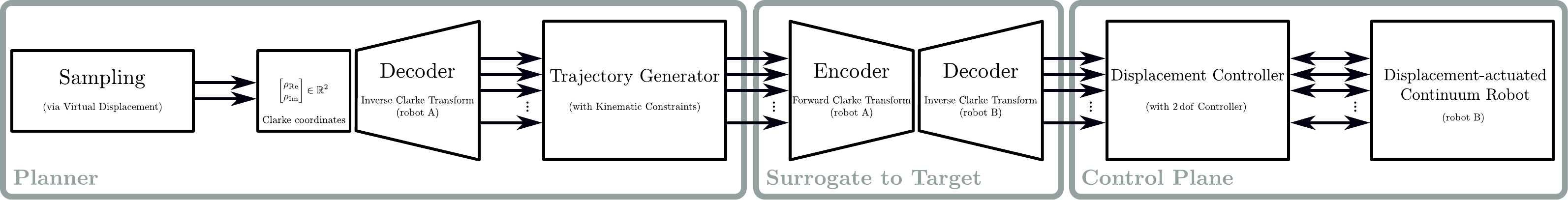}
    \vspace*{-1.75em}
    \caption{
        Workflow of the evaluation.
        }
    \label{fig:pipline}
    \vspace*{-1.5em}
\end{figure*}

We evaluate our proposed methods on a control problem in simulation.
For this, we state a two-staged method to generate feasible joint values, generate trajectories, and synthesize a PD controller scheme.
Figure~\ref{fig:pipline} illustrates the workflow.

For the evaluation, we consider five different displacement-actuated continuum robots with one type-0 segment, \textit{i.e.}, each continuum robot has a single segment with a fixed length.
The segment length is $l = \SI{0.1}{m}$ for all continuum robots, whereas the distance $d_i$ and angle $\psi_i$ are different.
Furthermore, the number of joints $n$ for each target continuum robot is different.
All kinematic design parameters are listed in Table~\ref{tab:design_parameters_evaluation}.

\begin{table*}
    \vspace*{0.75em}
    \caption{
        Kinematic design parameters of each considered displacement-actuated continuum robot.
    }
    \label{tab:design_parameters_evaluation}
    \vspace*{-0.75em}
    \centering
    \begin{tabular}{r r rr rr r} 
        \toprule
        robot & $n$ & $\psi_i$ in \SI{}{rad} & asymmetric $\psi_i$ & $d_i$ in \SI{}{mm} & non-constant $d_i$ & $l$ in \SI{}{m}\\
		\cmidrule(r){1-1}
		\cmidrule(lr){2-2}
		\cmidrule(lr){3-3}
		\cmidrule(lr){4-4}
		\cmidrule(lr){5-5}
		\cmidrule(lr){6-6}
		\cmidrule(l){7-7}
        \textit{robot\_0} & \num{3} & $2\pi\left[0,\ 1/3,\ 2/3\right]$ & \checkNo & $\left[10,\ 10,\ 10\right]$ & \checkNo & \num{0.1} \\[0.5em]
        \textit{robot\_A} & \num{4} & $2\pi\left[0,\ 0.25,\ 0.5,\ 0.75\right]$ & \checkNo & $\left[10,\ 10,\ 10,\ 10\right]$ & \checkNo & \num{0.1} \\[0.5em]
        \textit{robot\_B} & \num{3} & $2\pi\left[0,\ 1/3,\ 2/3\right]$ & \checkNo & $\left[10,\ 7,\ 5\right]$ & \checkYes & \num{0.1} \\[0.5em]
        \textit{robot\_C} & \num{5} & $2\pi\left[0,\ 0.2,\ 0.4,\ 0.6,\ 0.8\right]$ & \checkNo & $\left[10,\ 8.7,\ 5,\ 9.5,\ 6.5\right]$ & \checkYes & \num{0.1} \\[0.5em]
        \textit{robot\_D} & \num{7} & $2\pi\left[0.05,\ 0.18,\ 0.51,\ 0.63,\ 0.76,\ 0.87,\ 0.91\right]$ & \checkYes & $\left[10,\ 1,\ 8.7,\ 5,\ 5.6,\ 9.5,\ 6.5\right]$ & \checkYes & \num{0.1} \\
        \bottomrule
    \end{tabular}
    \vspace*{-1em}
\end{table*}

\subsection{Feasible Joint Values}

One of the used robot designs is a surrogate robot denoted by \textit{robot\_0}.
To sample $m + 1$ set of $n_{(\text{robot\_0})}$ feasible joint values, we use a two-stage method.

First, sample Clarke coordinates utilizing a direct sampling method \cite{GrassmannSenykBurgner-Kahrs_arXiv_2024}.
This rejection-free sampling method is vectorized.
Using the index notation, it can be expressed by
\begin{align}
    \begin{pmatrix}
        \rho_{\text{Re}, \mathcal{U}}^{\left(i\right)}\\[0.5em]
        \rho_{\text{Im}, \mathcal{U}}^{\left(i\right)}
    \end{pmatrix}_i
    =&
    \begin{pmatrix}
        L_\mathcal{U}^{\left(i\right)}\cos\!\left(\theta_\mathcal{U}^{\left(i\right)}\right)\\[0.5em]
        L_\mathcal{U}^{\left(i\right)}\sin\!\left(\theta_\mathcal{U}^{\left(i\right)}\right)
    \end{pmatrix}_i
    \in
    \mathbb{R}^{2 \times m+1}
    ,
    \nonumber
\end{align}
where all magnitude $L_\mathcal{U}^{\left(i\right)}$ and all angle $\theta_\mathcal{U}^{\left(i\right)}$ are sampled via
\begin{align}
    L_\mathcal{U}
    =&
    \,
    \pi d_{(\text{robot\_0})}\sqrt{\mathcal{U}^{m+1}\left[0; 1\right]}
    \quad\text{and}
    \nonumber
    \\
    \theta_\mathcal{U}
    =&
    \,
    \pi\mathcal{U}^{m+1}\left[-1; 1\right)
    ,
    \nonumber
\end{align}
respectively.
The distance $d_{(\text{robot\_0})}$ is $\SI{0.01}{m}$ as stated in Table~\ref{tab:design_parameters_evaluation}.
The expression $\sqrt{\mathcal{U}^{m+1}\left[0; 1\right]}$ samples $m+1$ values from an uniform distribution $\mathcal{U}$ with the interval $\left[0; 1\right]$ and, afterwards, the square root is computed.
This leads to a uniformly distributed disk \cite{GrassmannSenykBurgner-Kahrs_ICRA_2024}.
Note that, assuming constant curvature assumption, $\max L_\mathcal{U} = \pi d_{\left(\text{robot\_0}\right)}$ corresponds to a half-circle.
This formulation allows specifying the maximal value without defining the maximal curvature and using segment length $l$.

Second, transform all sampled Clarke coordinates using \eqref{eq:inverse}, which can be vectorized as well.
This step is illustrated as the decoder in Fig.~\ref{fig:pipline}.

\subsection{Trajectory Generation}

To provide smooth trajectories for $n$ displacements, we adapted a $\mathcal{C}^4$-smooth trajectory generator \cite{GrassmannBurgner-Kahrs_RAL_2019} for via poses capable of respecting kinematic limits, \textit{i.e.}, considering the maximum velocity and maximum acceleration.
Here, we use the fact that $\rhovec \subset \mathbb{R}^n$ can be treated as $n\,\SI{}{dof}$ position in Euclidean space.

For $m-1$ intermediate points, one start point, and a goal point, we define $m\times n$ trajectories for $n$ displacements.
Each trajectory is composed of three phases for the trapezoidal-like velocity profile.
They are denoted by \textit{lo}, \textit{cr}, and \textit{sd} for \textit{lift-off}, \textit{cruise}, and \textit{set-down} phase, respectively.
The $m\times n$ trajectories are blended into $n$ trajectories for each displacement $\rho_i$.
We kindly refer to \cite{GrassmannBurgner-Kahrs_RAL_2019} for details on the blending.

In contrast to \cite{GrassmannBurgner-Kahrs_RAL_2019}, the used $j\textsuperscript{th}$ trajectory state $\mathcal{T}^{j}_i$ of the $m$ trajectories of the $i\textsuperscript{th}$ displacement is defined as
\begin{align}
    \mathcal{T}^{j}_i = \left(\Delta\rho_i^j,\ v_i^j,\ a_i^j,\ d_i^j,\ t_{\mathrm{lo}, i}^j,\ t_{\mathrm{cr},\ i}^j,\ t_{\mathrm{sd}, i}^j,\ t_{\text{enb}, i}^j \right),
    \label{eq:trajectory_state_Chapter_Clarke}
\end{align}
where the description of each variable is listed in Table~\ref{tab:trajectory_state_Chapter_Clarke}.
An advantage of \eqref{eq:trajectory_state_Chapter_Clarke} is the tracking of each duration, which simplifies the synchronization between the $n$ trajectories, reduces re-computation of quantities, and avoids checking of zero divisions. 
Related to the time-memory trade-off in computer science, this comes at the expense of a slightly larger trajectory state formulation, \textit{i.e.}, eight variables in \eqref{eq:trajectory_state_Chapter_Clarke} instead of five variable, \textit{cf.} \cite{GrassmannBurgner-Kahrs_RAL_2019}.

\begin{table}
	\centering
	\caption[Trajectory state representation.]{
        Trajectory state representation.
    }
	\label{tab:trajectory_state_Chapter_Clarke}
    \vspace*{-.75em}
	\begin{tabular}{@{} r  p{6.5cm} @{}}
		\toprule
		\multicolumn{1}{N}{Variable}
		& \multicolumn{1}{N}{Description of the variable in $\mathcal{T}^{j}_i$}\\
		\cmidrule(r){1-1}
		\cmidrule(l){2-2}
	    $\Delta\rho_i^j$ & Distance to overcome. It is define as $\Delta\rho_i^j = \rho_i^j - \rho_i^{j-1}$\\[.5em]
	    $\rho_i^{j-1}$ & Start point of the displacement\\[.5em]
	    $\rho_i^j$ & Goal point of the displacement\\[.5em]
	    $v_i^j$ & Maximum velocity of the velocity profile\\[.5em]
	    $a_i^j$ & Maximum acceleration of the lift-off phase\\[.5em]
	    $d_i^j$ & Maximum deacceleartion of the set-down phase\\[.5em]
	    $t_{\mathrm{lo}, i}^j$ & Duration for the lift-off phase\\[.5em]
	    $t_{\mathrm{cr}, i}^j$ & Duration for the curse phase\\[.5em]
	    $t_{\mathrm{sd}, i}^j$ & Duration for the set-down phase\\[.5em]
	    $t_{\mathrm{enb}, i}^j$ & Switching time at which the $j\textsuperscript{th}$ trajectory starts and the blending into the $(j+1)\textsuperscript{th}$ trajectory is happening\\
		\bottomrule
	\end{tabular}
\end{table}

\begin{figure}
    \centering
    \includegraphics[width=0.9\columnwidth, trim={0, 20, 25, 10}, clip]{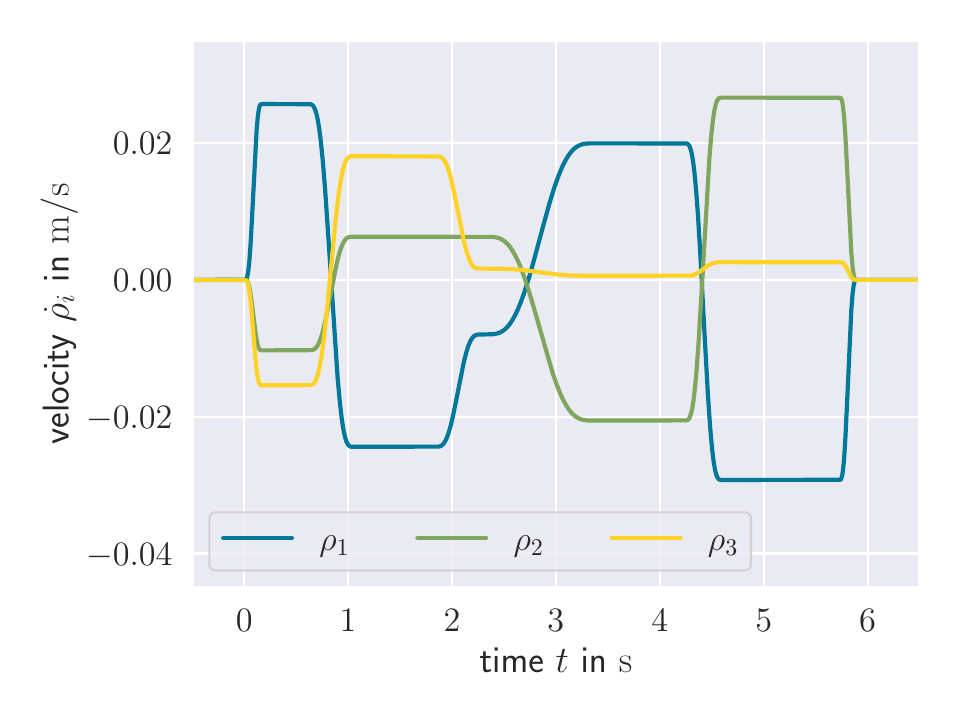}
    \vspace*{-1em}
    \caption{
        Velocity profile of 
        the surrogate robot with $n = n_{(\text{robot\_0})} = 3$.
        The trapezoidal-like velocity profiles and blending of the velocity profiles of each displacement are $\mathcal{C}^4$-smooth.
        }
    \label{fig:velocity_profile_robot_0}
\end{figure}

For the evaluation, the kinematic constraints are set to $v_i^j = \SI{0.01\pi}{m/s}$, $a_i^j = \SI{0.125\pi}{m/s^2}$, and $d_i^j = \SI{0.125\pi}{m/s^2}$ for all $i$ and $j$.
The generated paths consist of $m = 5$ blended velocity profiles.
Figure~\ref{fig:velocity_profile_robot_0} shows the resulted velocity profiles for the surrogate robot.

\subsection{Control of Joints}

\begin{figure}[thb]
    \centering
    \includegraphics[width=\columnwidth]{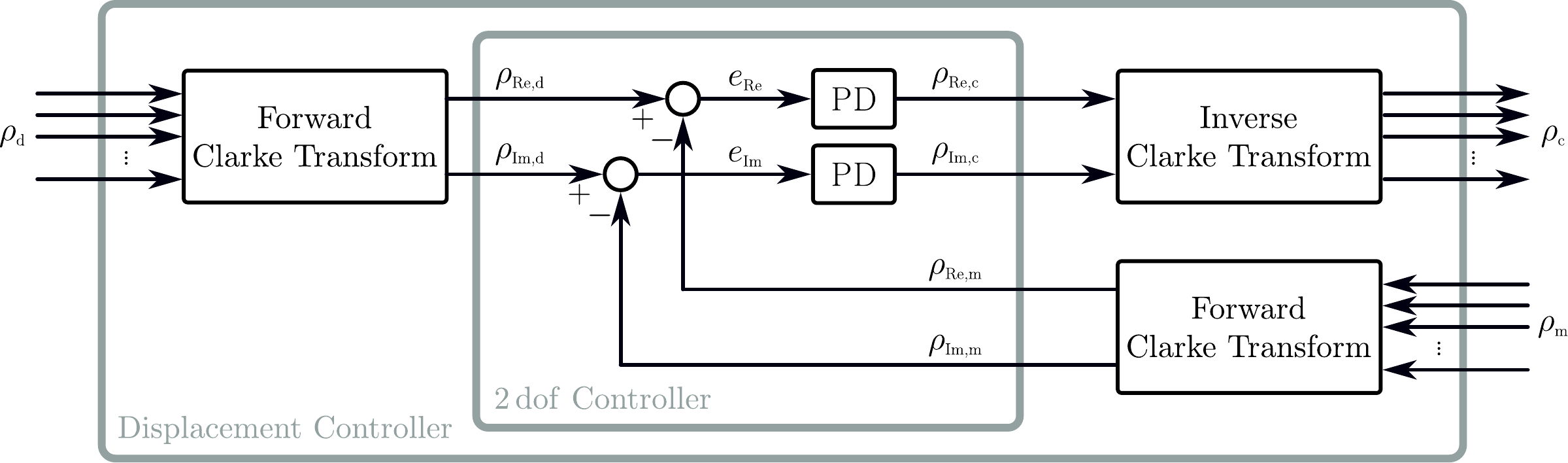}
    \vspace*{-1.5em}
    \caption{
        Control of Clarke Coordinates. 
        Using the Clarke transform, only two simple PD controllers are necessary to control $n$ displacement-actuated joints.
        }
    \vspace*{-0.5em}
    \label{fig:controller}
\end{figure}

The setup is similar to \cite{GrassmannSenykBurgner-Kahrs_arXiv_2024}, where the control frequency is \SI{1}{kHz}, each actuator is modelled as an independent first-order
proportional delay element (PT\textsubscript{1}) system with a time constant equal to $\SI{250}{ms}$, and the additive measurement noise is drawn from a uniform distribution $\mathcal{U}^n\left[-\epsilon, \epsilon\right]$ with $\epsilon = \SI{2.5}{mm}$ every \SI{1}{ms}.
Here, all PD control gains are $\boldsymbol{K}_\mathrm{p} = 75$ and $\boldsymbol{K}_\mathrm{d} = 0.0015$, which are set manually without a specific heuristic.
The controller scheme is illustrated in Fig.~\ref{fig:controller}.

\begin{figure}
    \centering
    \hfill
    \includegraphics[height=2.95cm, trim={0, 60, 25, 10}, clip]{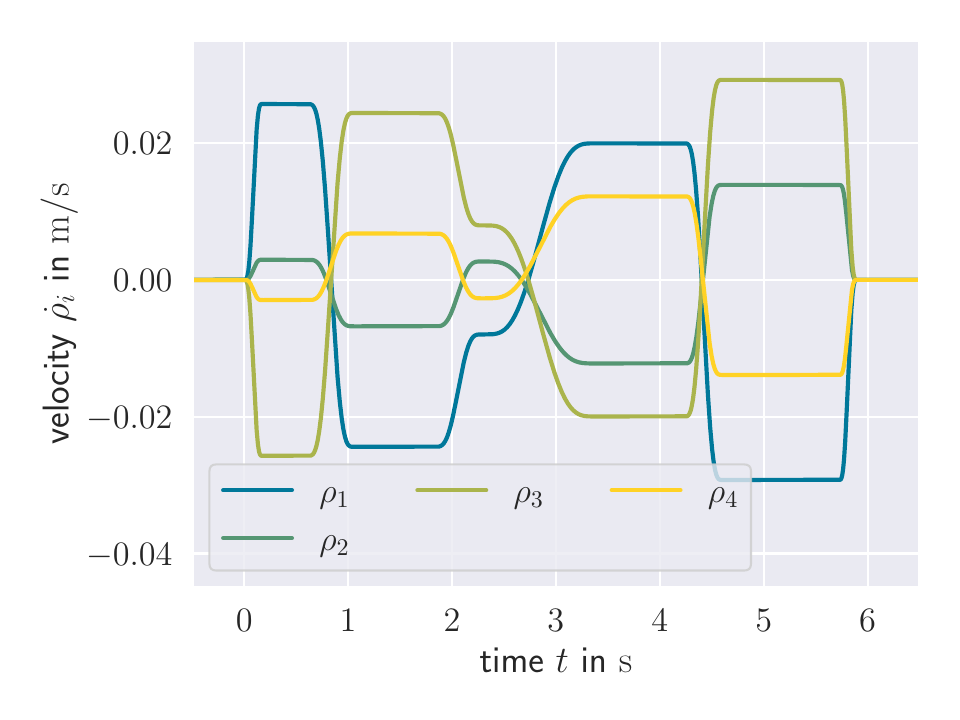}
    \includegraphics[height=2.95cm, trim={85, 60, 25, 10}, clip]{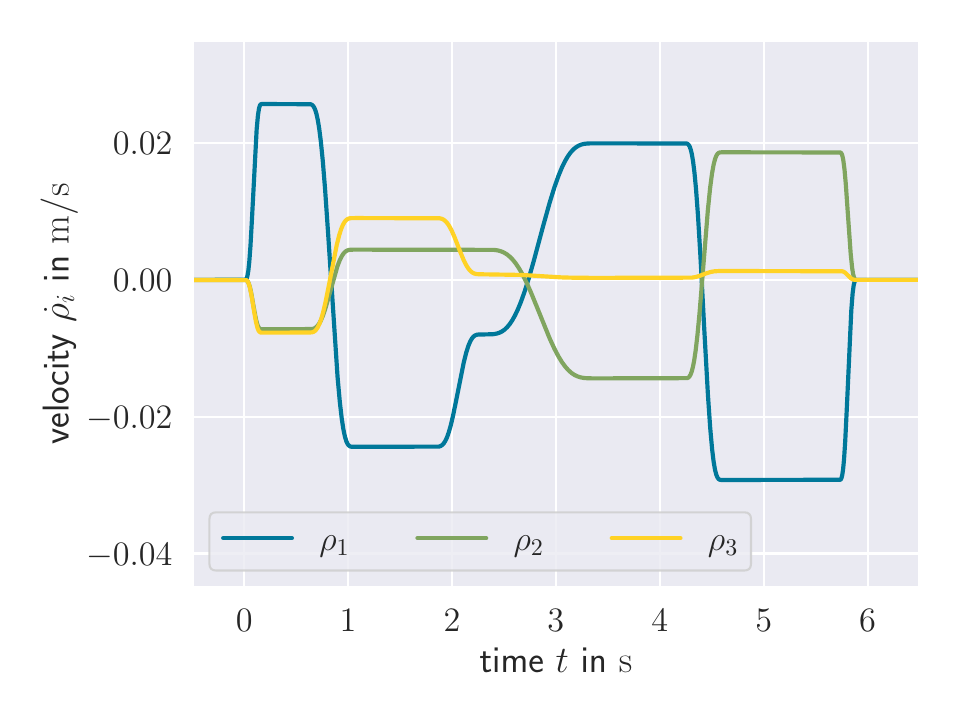}
    \\
    \hfill
    \includegraphics[height=3.3775cm, trim={0, 20, 25, 10}, clip]{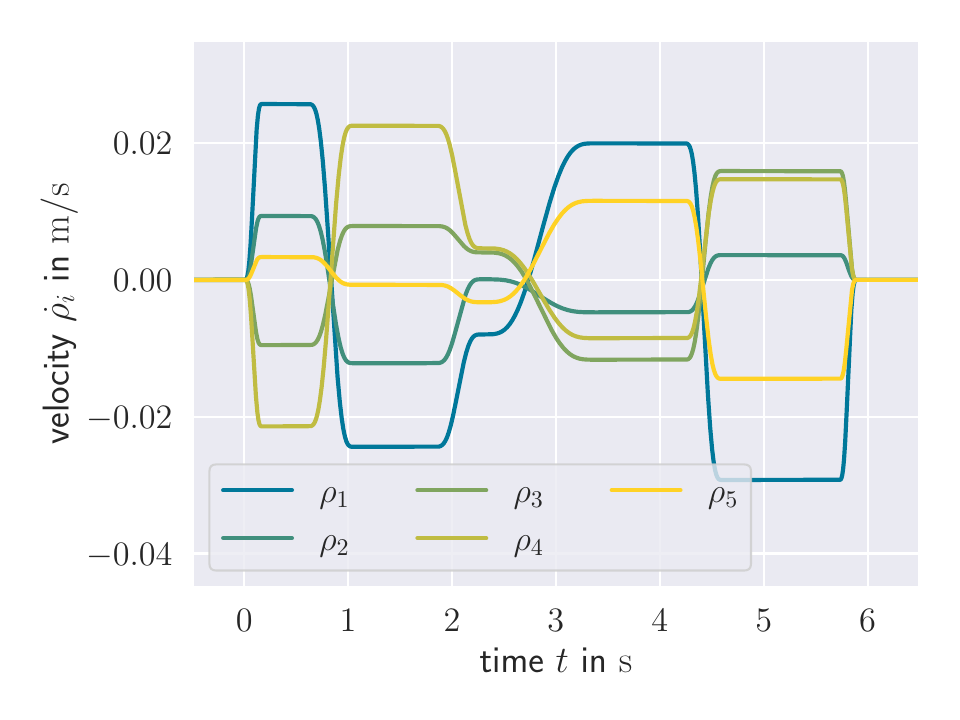}
    \includegraphics[height=3.3775cm, trim={85, 20, 25, 10}, clip]{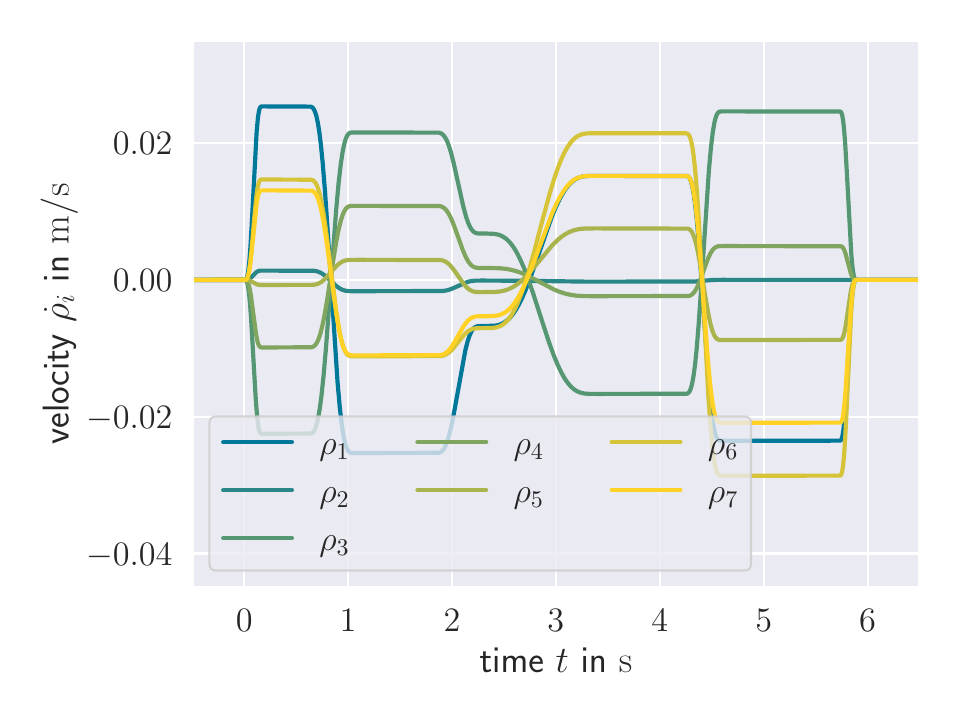}
    \caption{
        Transformed velocity profiles as the output of a used encoder-decoder architecture.
        The number of displacements unambiguously identifies the corresponding target robot listed in Table~\ref{tab:design_parameters_evaluation}.
        As can be seen, the kinematic constraint on the velocity, \textit{i.e.}, $v_i^j = \SI{0.01\pi}{m/s} \approx \SI{0.03}{m/s}$, is fulfilled.
        }
    \label{fig:velocity_profiles_traget_robot}
    \vspace*{-0.75em}
\end{figure}

The desired displacements $\rhovec_\mathrm{d}$ of the target robot for each time step are provided by the encoder-decoder architecture, see Fig.~\ref{fig:pipline}.
The encoder-decoder architecture illustrated in Fig.~\ref{fig:encoder-decoder} transforms the trajectory generated for surrogate robot \textit{robot\_0} to the respective target robot, \textit{e.g.}, \textit{robot\_A} or \textit{robot\_D}.
Therefore, the encoder part is constant through the evaluation, whereas the decoder part depends on the target robot listed in Table~\ref{tab:design_parameters_evaluation}.
The transformed velocity profiles for each target robot are shown in Fig.~\ref{fig:velocity_profiles_traget_robot}.

\subsection{Results}

\begin{figure*}
    \centering
    \vspace*{0.5em}
    \hfill
    \begin{overpic}[height=4.15cm, trim={0, 60, 25, 10}, clip]{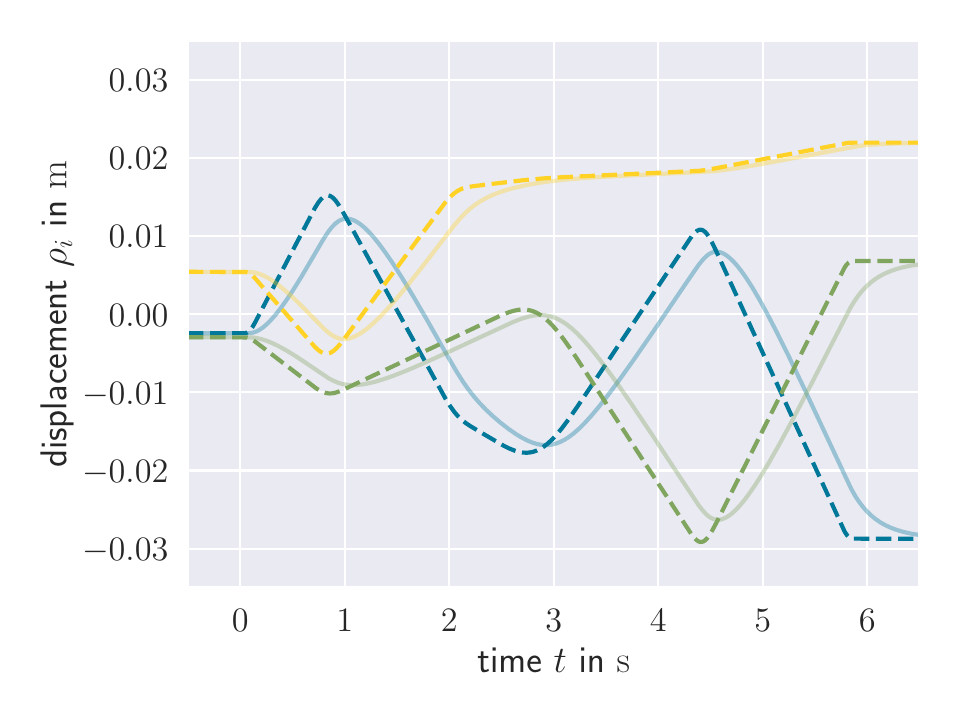}
        \put(23,63){Open-loop behavior without noise}
        \put(-5,24){\rotatebox{90}{\textit{robot\_0}}}
        \put(27, 6){\small $n = 3$}
        \put(24,50){\small $\rho_1$\vector(1,-1){5}}
        \put(61,6){\small $\rho_2$\vector(1,1){5}}
        \put(55,53){\small $\rho_3$\vector(1,-1){5}}
    \end{overpic}
    \begin{overpic}[height=4.15cm, trim={85, 60, 25, 10}, clip]{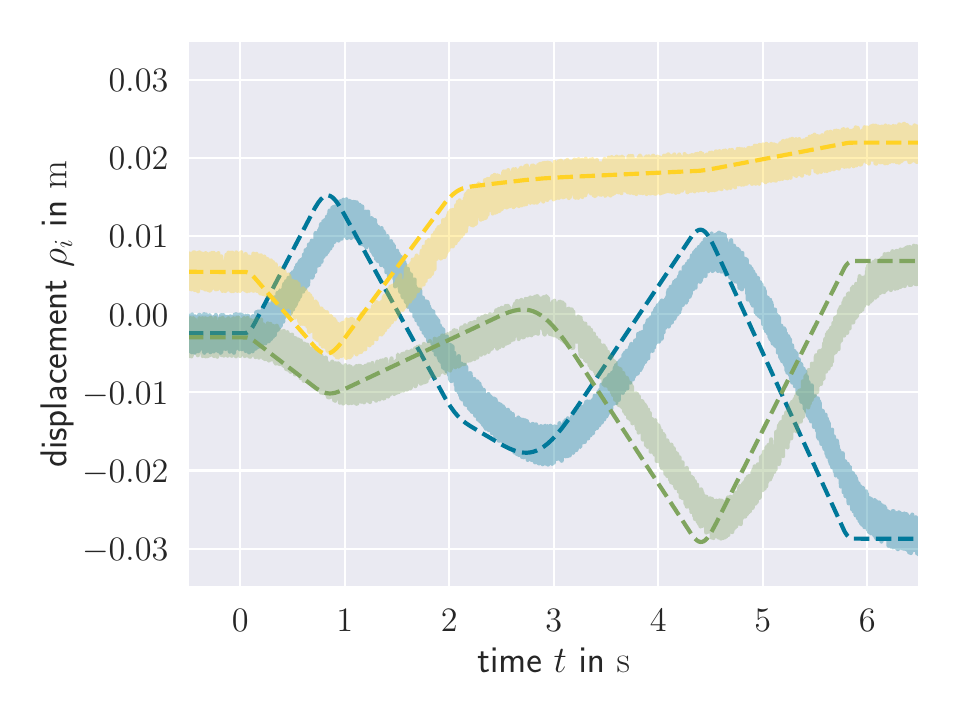}
        \put(9.5,78.25){Open-loop behavior with noise}
    \end{overpic}
    \begin{overpic}[height=4.15cm, trim={85, 60, 25, 10}, clip]{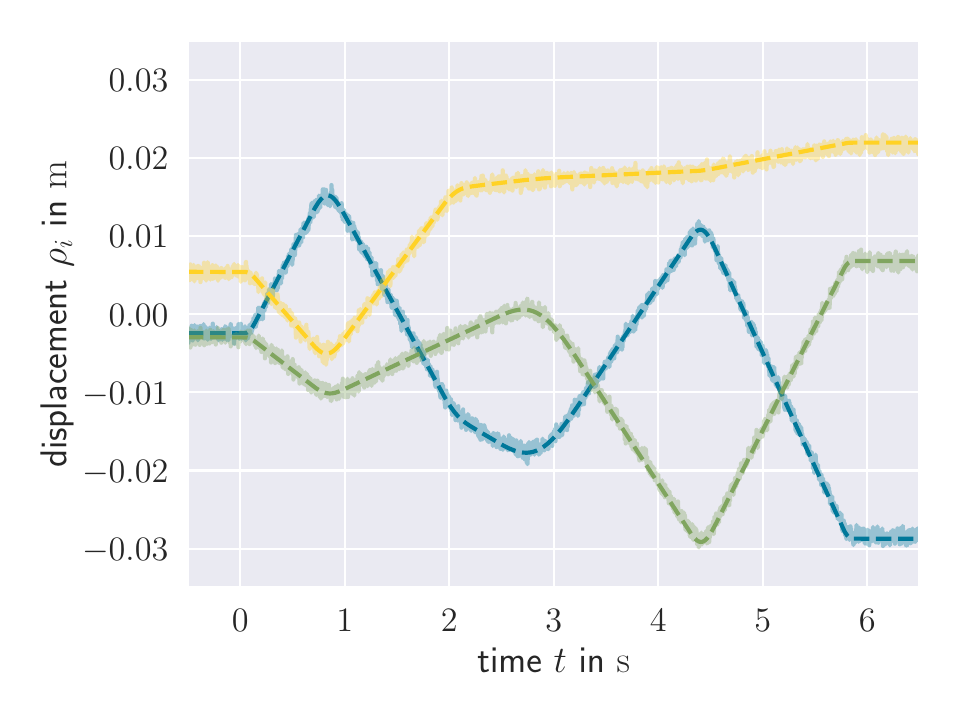}
        \put(22.5,78.25){Closed-loop behavior}
        \put(3, 3){\small $d_1 = d_2 = d_3 = \SI{10}{mm}$}
    \end{overpic}
    \\
    \hfill
    \begin{overpic}[height=4.15cm, trim={0, 60, 25, 10}, clip]{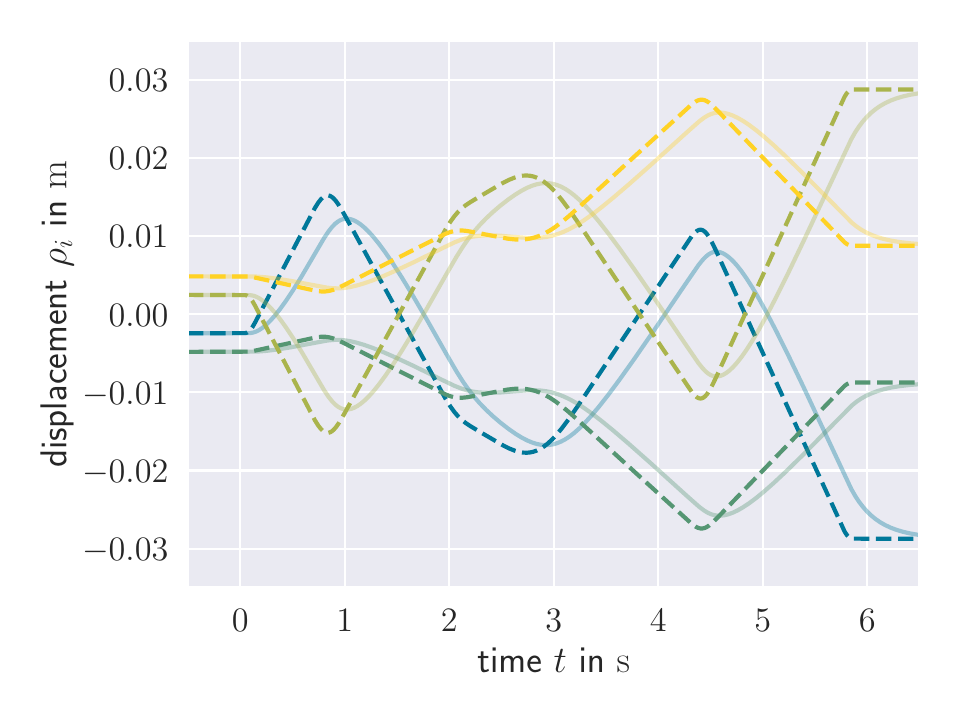}
        \put(-5,24){\rotatebox{90}{\textit{robot\_A}}}
        \put(27, 6){\small $n = 4$}
        \put(24,50){\small $\rho_1$\vector(1,-1){5}}
        \put(61,6){\small $\rho_2$\vector(1,1){5}}
        \put(42,50){\small $\rho_3$\vector(1,-1){5}}
        \put(62,57){\small $\rho_4$\vector(1,-1){5}}
    \end{overpic}
    \includegraphics[height=4.15cm, trim={85, 60, 25, 10}, clip]{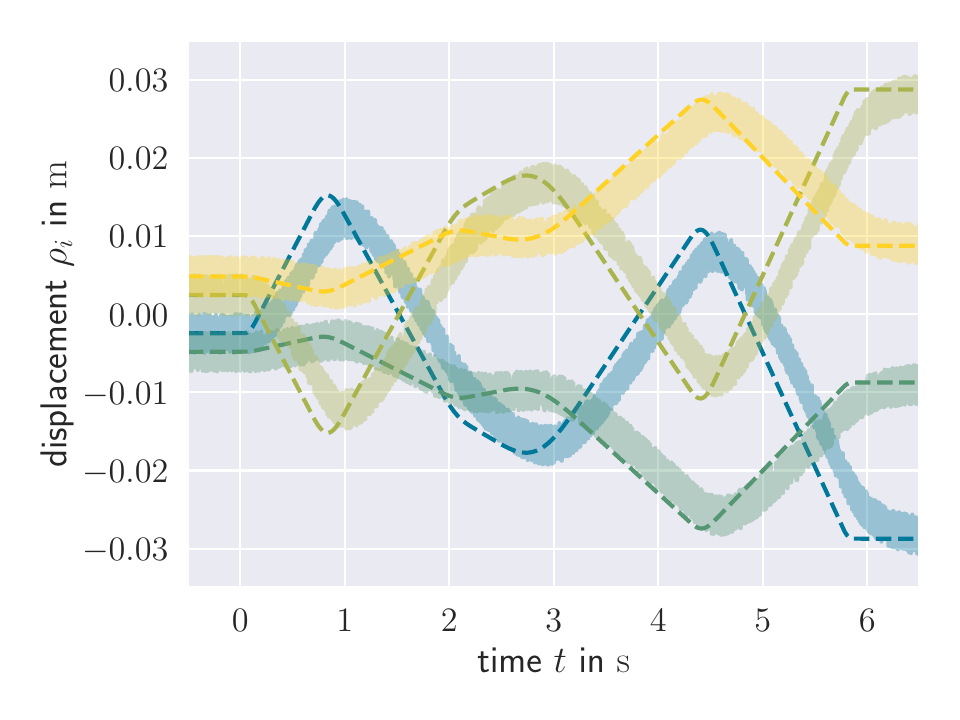}
    \begin{overpic}[height=4.15cm, trim={85, 60, 25, 10}, clip]{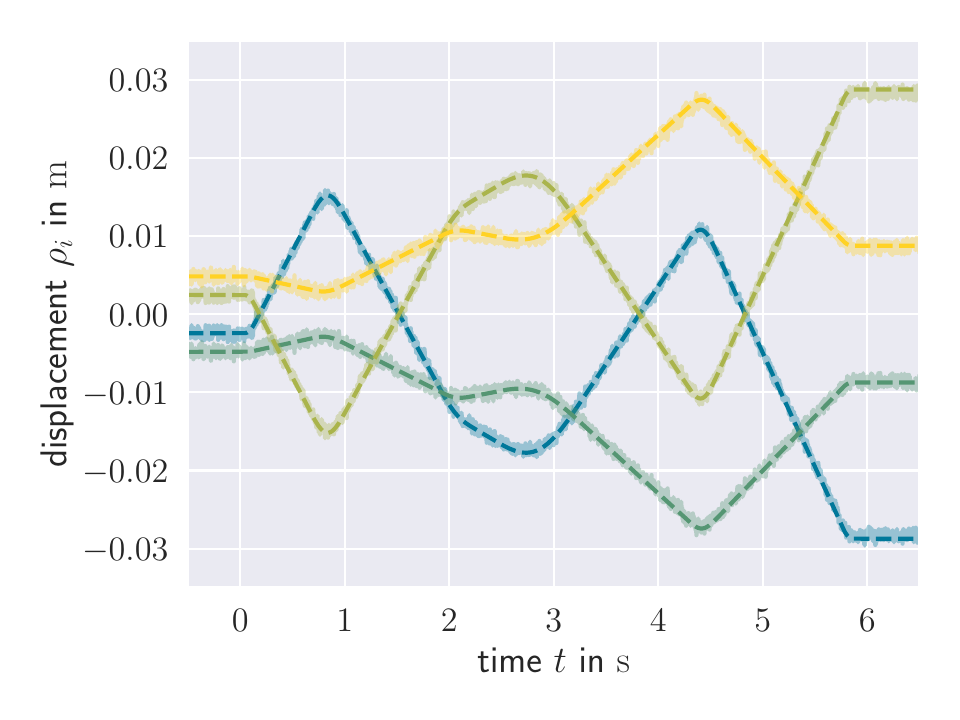}
        \put(3, 3){\small $d_1 = d_2 = d_3 = d_4 = \SI{10}{mm}$}
    \end{overpic}
    \\
    \hfill
    \begin{overpic}[height=4.15cm, trim={0, 60, 25, 10}, clip]{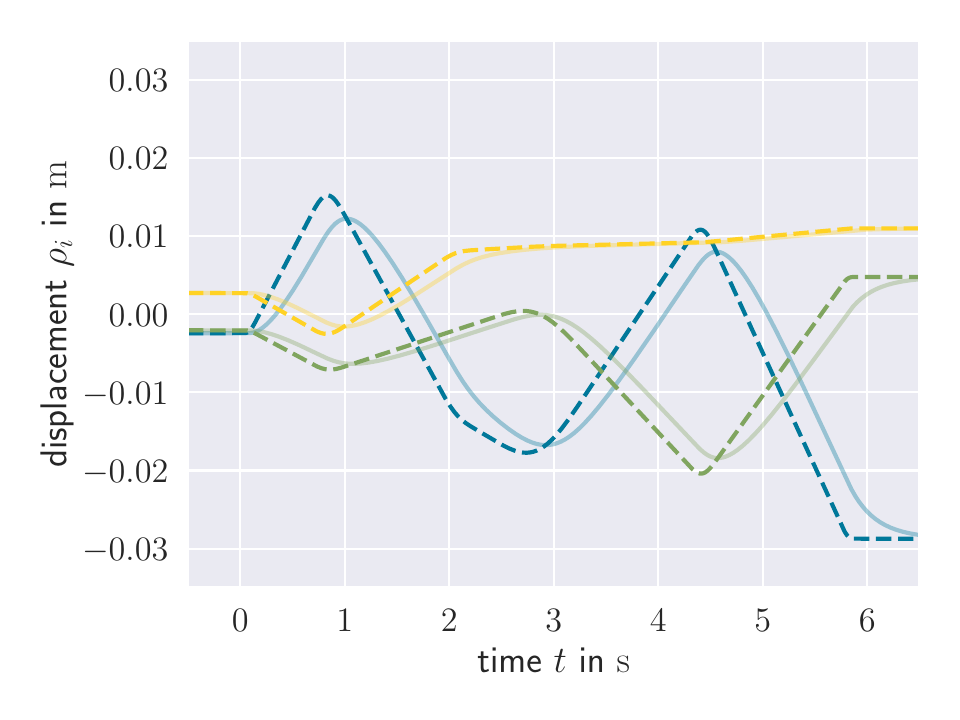}
        \put(-5,24){\rotatebox{90}{\textit{robot\_B}}}
        \put(27, 6){\small $n = 3$}
        \put(24,50){\small $\rho_1$\vector(1,-1){5}}
        \put(65,6){\small $\rho_2$\vector(1,1){5}}
        \put(50,45){\small $\rho_3$\vector(1,-1){5}}
    \end{overpic}
    \includegraphics[height=4.15cm, trim={85, 60, 25, 10}, clip]{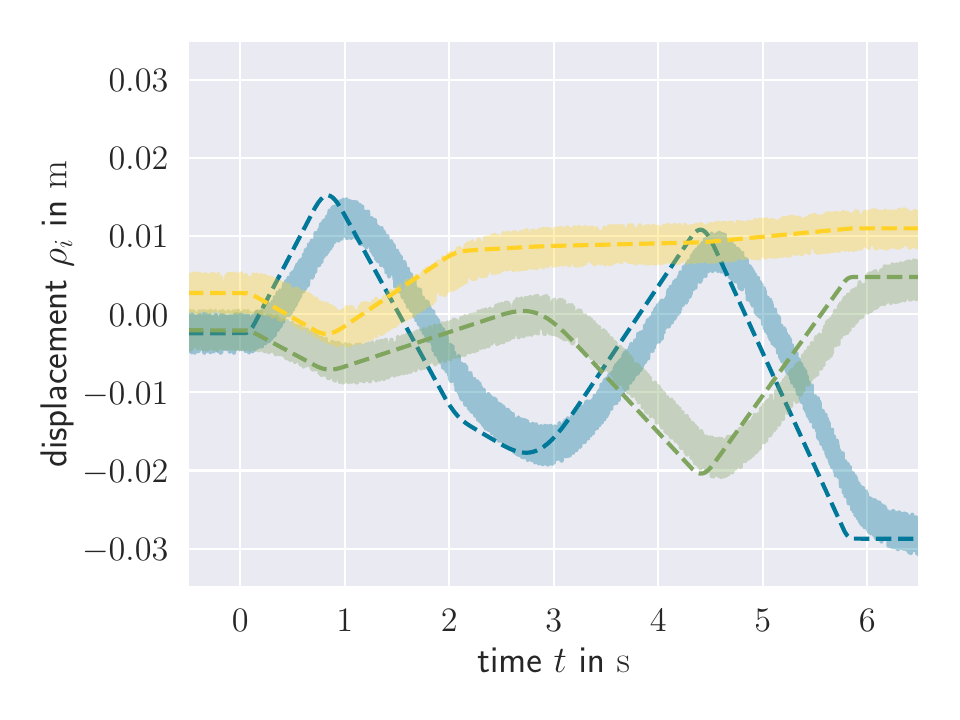}
    \begin{overpic}[height=4.15cm, trim={85, 60, 25, 10}, clip]{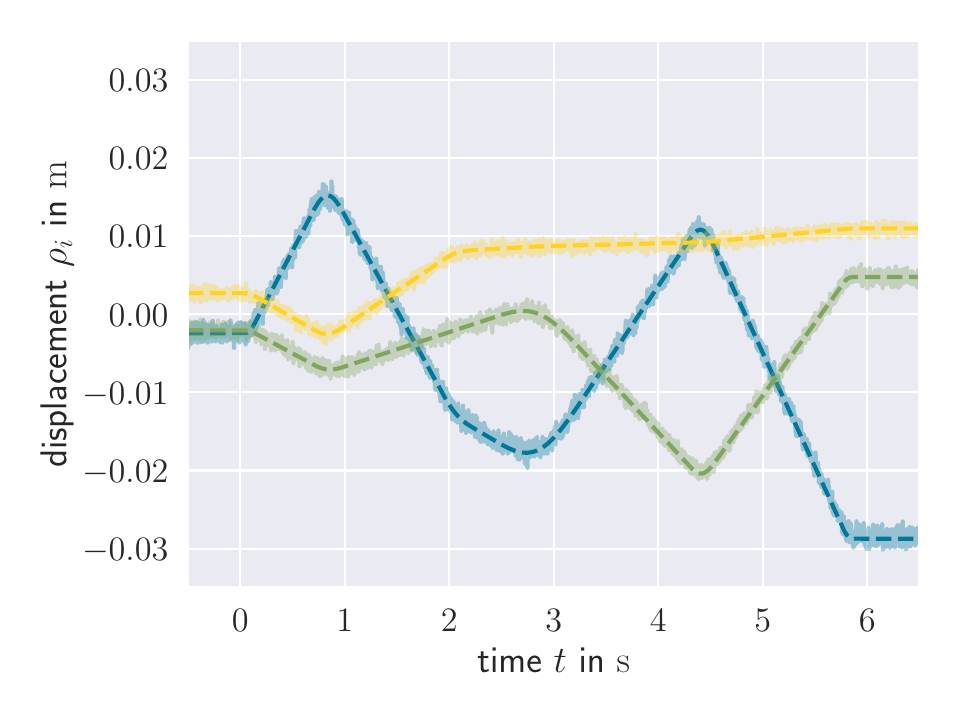}
        \put(28,66){\small $d_1 = \SI{10}{mm}$}
            \put(27,65){\small \vector(-1,-1){5}}
        \put(53,6){\small $d_2 = \SI{7}{mm}$}
            \put(58,12){\vector(1,1){7}}
        \put(37,57){\small $d_3 = \SI{5}{mm}$}
            \put(43,55){\small\vector(1,-1){5}}
    \end{overpic}
    \\
    \hfill
    \begin{overpic}[height=4.15cm, trim={0, 60, 25, 10}, clip]{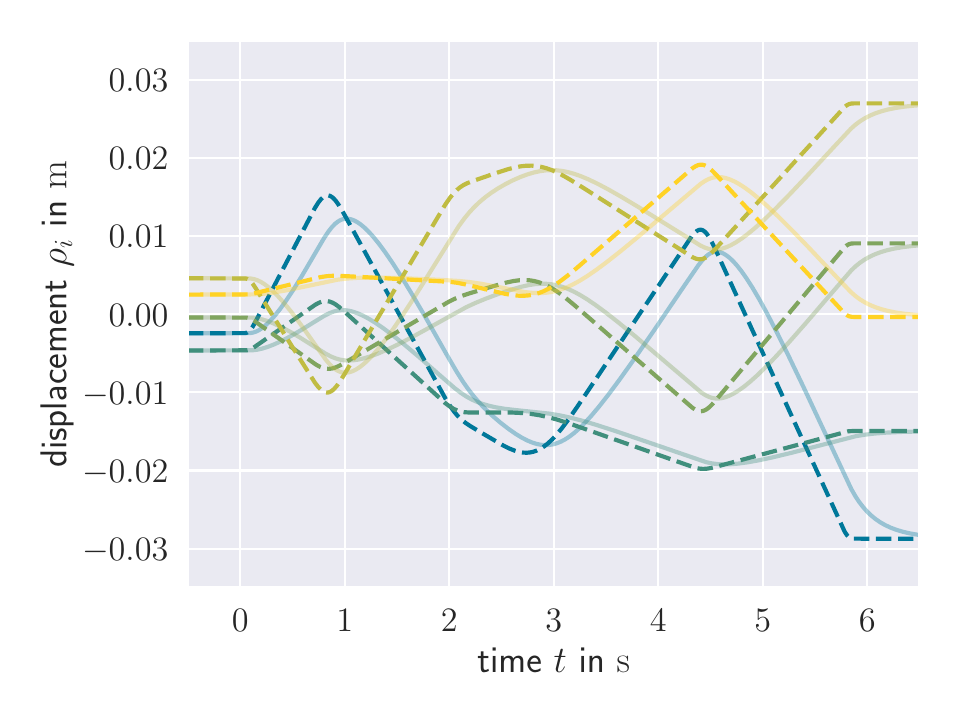}
        \put(-5,24){\rotatebox{90}{\textit{robot\_C}}}
        \put(27, 6){\small $n = 5$}
        \put(24,50){\small $\rho_1$\vector(1,-1){5}}
        \put(62,9){\small $\rho_2$\vector(1,1){5}}
        \put(92,45){\small $\rho_3$\vector(0,-1){5}}
        \put(45,53){\small $\rho_4$\vector(1,-1){5}}
        \put(65,53){\small $\rho_5$\vector(1,-1){5}}
    \end{overpic}
    \includegraphics[height=4.15cm, trim={85, 60, 25, 10}, clip]{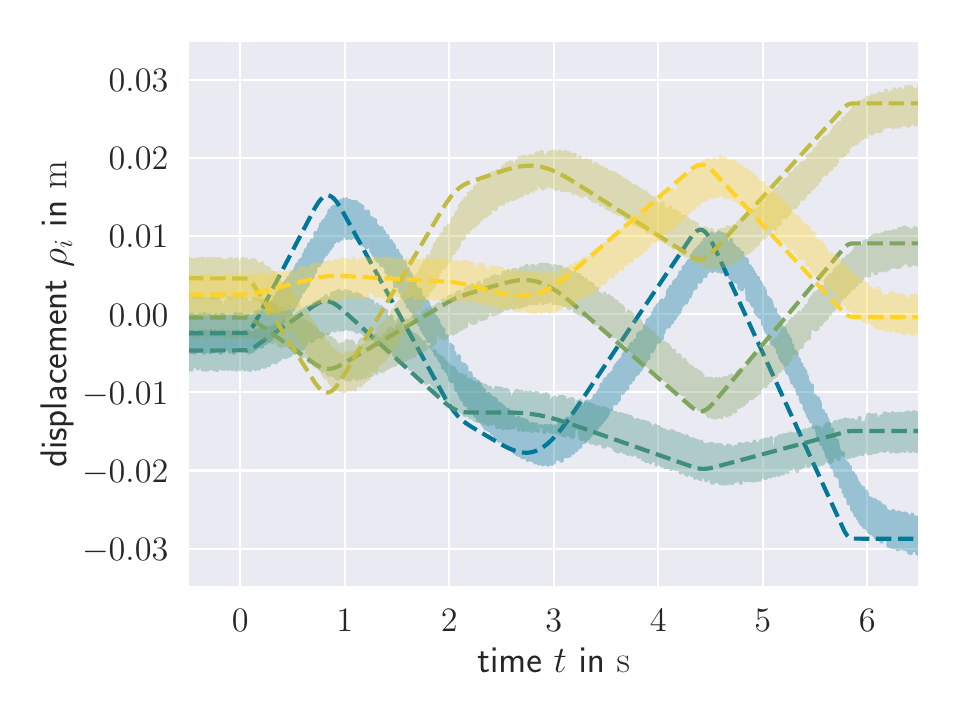}
    \begin{overpic}[height=4.15cm, trim={85, 60, 25, 10}, clip]{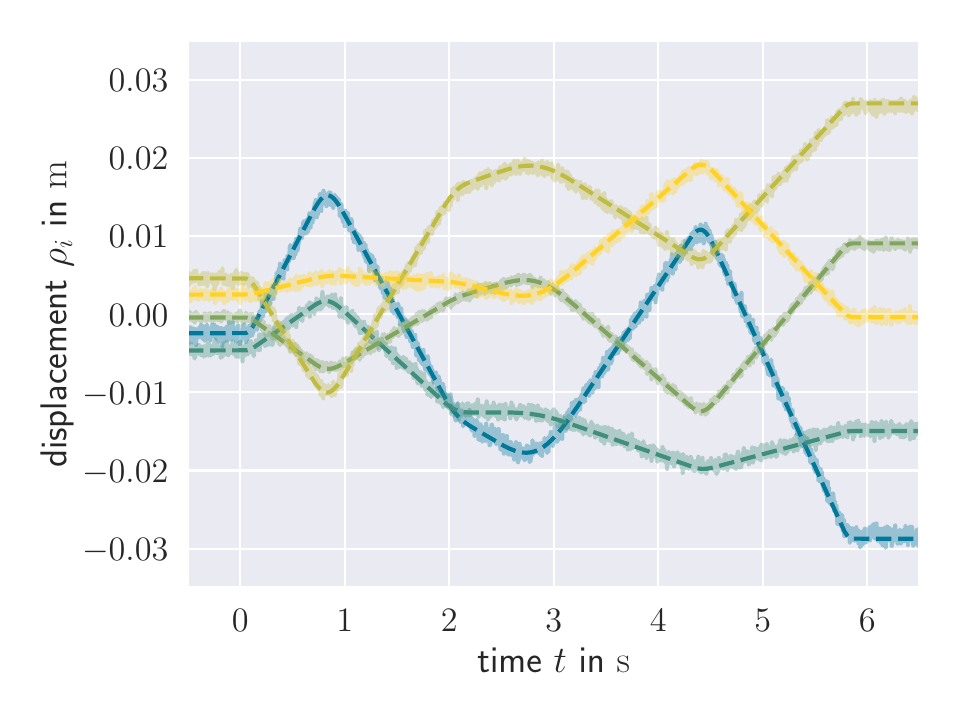}
        \put(28,66){\small $d_1 = \SI{10}{mm}$}
            \put(27,65){\small \vector(-1,-1.5){5}}
        \put(52,5){\small $d_3 = d_1 / 2$}
            \put(57,10){\small \vector(1.5,3){8.5}}
    \end{overpic}
    \\
    \hfill
    \begin{overpic}[height=4.75cm, trim={0, 20, 25, 10}, clip]{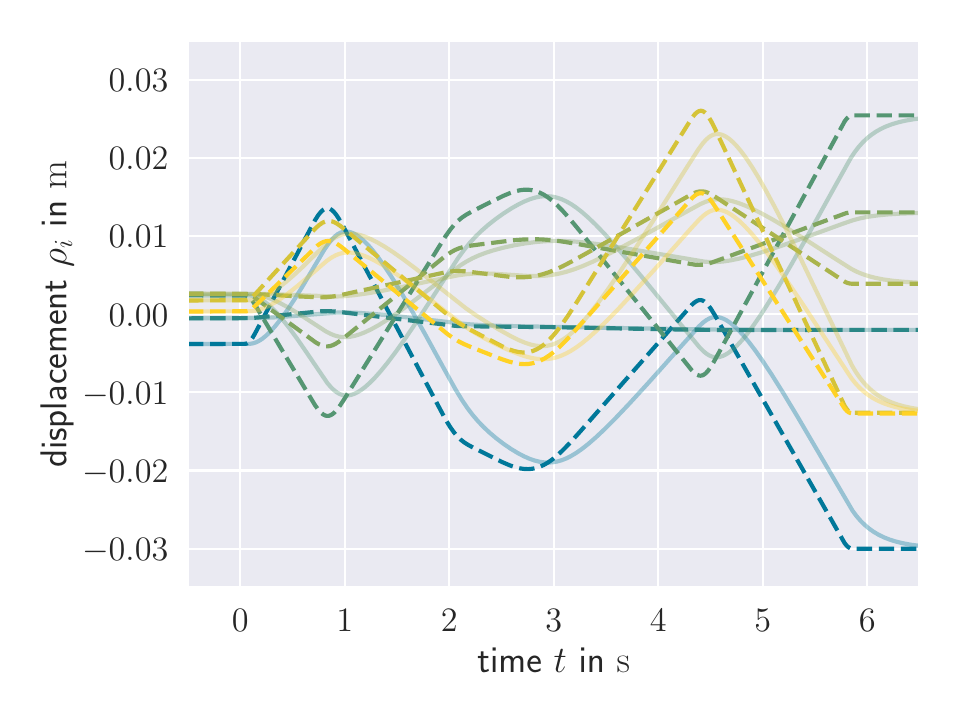}
        \put(-5,33.5){\rotatebox{90}{\textit{robot\_D}}}
        \put(27, 15){\small $n = 7$}
        \put(78.5,14.5){\small $\rho_1$\vector(1,1){5}}
        \put(61,32){\small $\rho_2$\vector(1,2){2.5}}
        \put(24,23){\small $\rho_3$\vector(1,1){5}}
        \put(94,56){\small $\rho_4$\vector(0,-1){4}}
        \put(94,48){\small $\rho_5$\vector(0,-1){4}}
        \put(64,66){\small $\rho_6$\vector(1,-1){5}}
        \put(54,29){\small $\rho_7$\vector(0,1.4){5}}
    \end{overpic}
    \includegraphics[height=4.75cm, trim={85, 20, 25, 10}, clip]{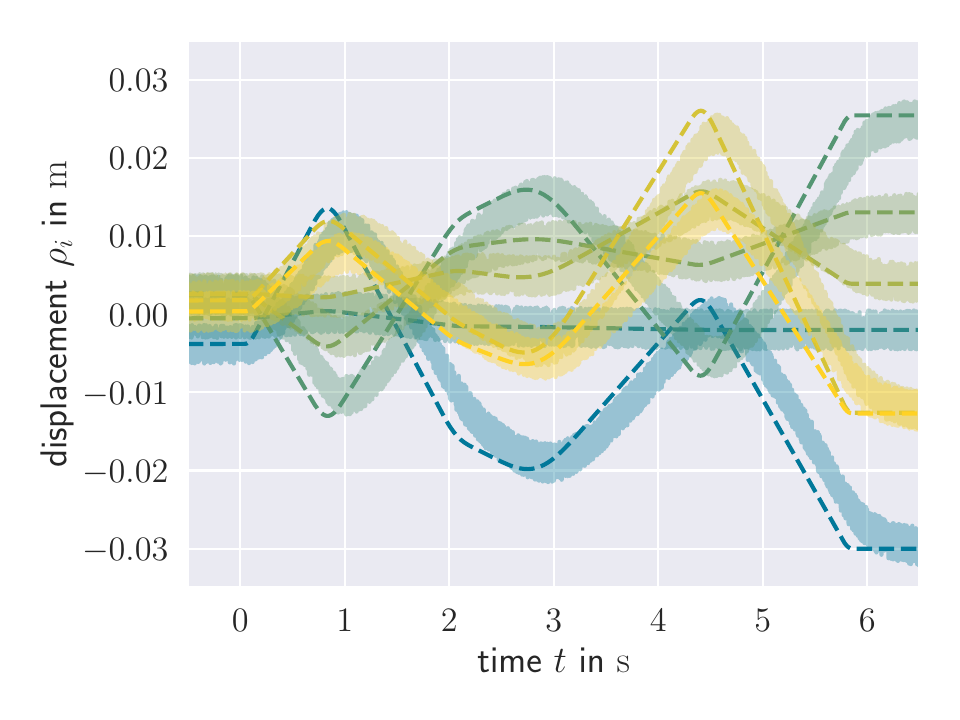}
    \begin{overpic}[height=4.75cm, trim={85, 20, 25, 10}, clip]{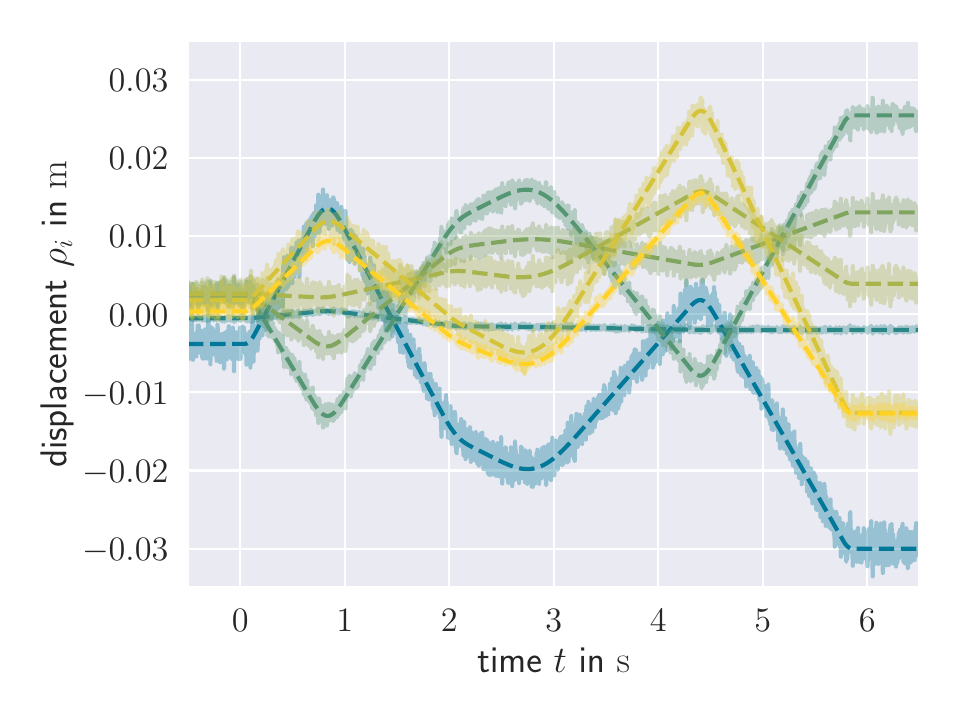}
        \put(3,18){\small $d_1 = \SI{10}{mm}$}
            \put(9,23){\small \vector(7,4){24}}
        \put(55,18){\small $d_2 = d_1 / 10$}
            \put(61,23){\small \vector(3,2){34}}
        \put(73.5,82){\small $d_4 = d_1 / 2$}
            \put(80,80){\small \vector(1,-1){13}}
    \end{overpic}
    \\[0em] 
    \vspace*{-0.75em}
    \caption{
        Displacement-control.
        A desired path is indicated by a dotted line, whereas a solid line is the corresponding system behavior.
        (left column) Desired path versus open-loop behavior of the noise-free PT\textsubscript{1} system.
        (middle column) Desired path versus measured displacement with noise.
        Due to the high frequency, the measurement noise appears to be a band.
        (right column) Desired path versus closed-loop behavior.
        (first row) surrogate robot \textit{robot\_0} with $n = 3$ symmetric joint location.
        (second row) target robot \textit{robot\_A} with $n = 4$ with symmetric joint location.
        (third row) target robot \textit{robot\_B} with $n = 3$ with non-constant distant $d_i$.
        (fourth row) target robot \textit{robot\_C} with $n = 5$ with non-constant distant $d_i$.
        (last row) target robot \textit{robot\_D} with $n = 7$ with non-constant distant $d_i$ and asymmetric $\psi_i$.
        }
    \label{fig:results_closed_loop_advance}
\end{figure*}

\begin{figure*}
    \centering
    \vspace*{1em}
    \hfill
    \begin{overpic}[height=4.15cm, trim={0, 60, 25, 10}, clip]{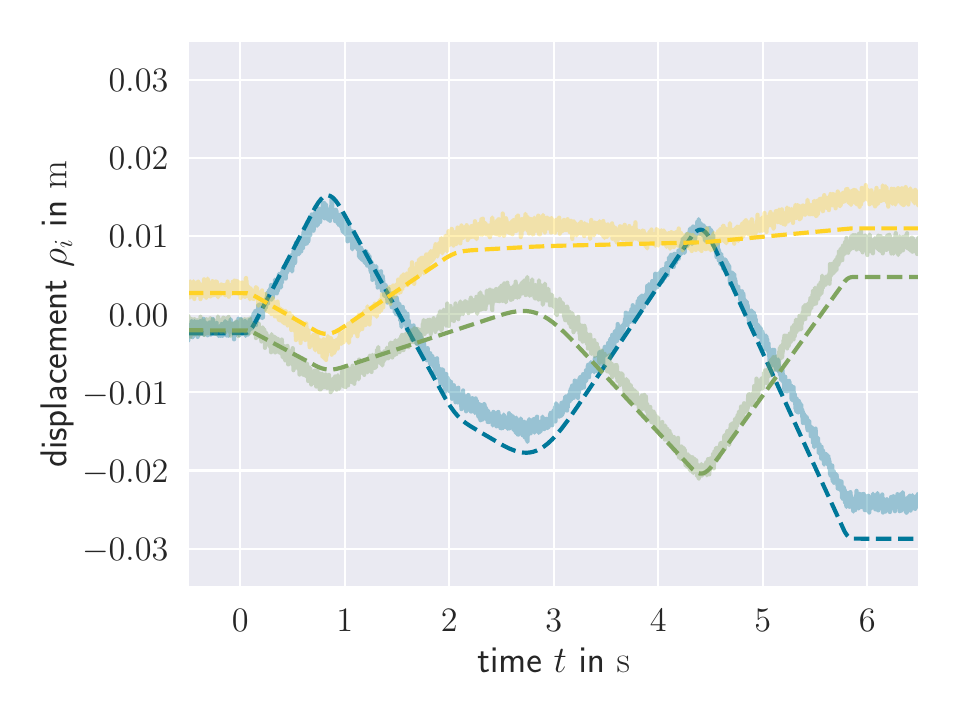}
        \put(53,63){\textit{robot\_B}}
    \end{overpic}
    \begin{overpic}[height=4.15cm, trim={85, 60, 25, 10}, clip]{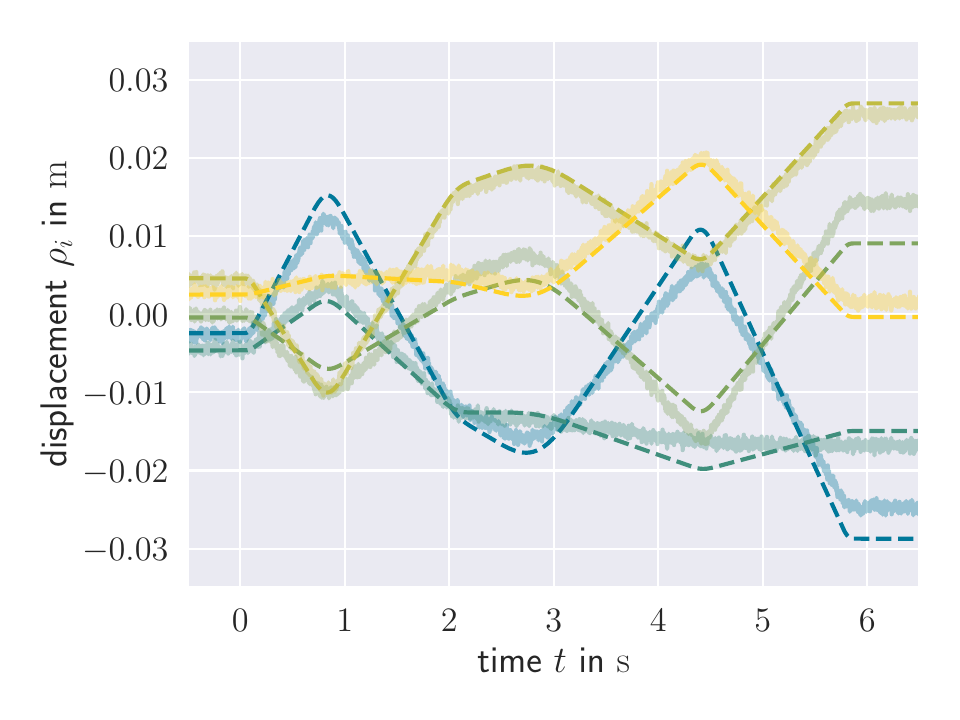}
        \put(42,78.25){\textit{robot\_C}}
    \end{overpic}
    \begin{overpic}[height=4.15cm, trim={85, 60, 25, 10}, clip]{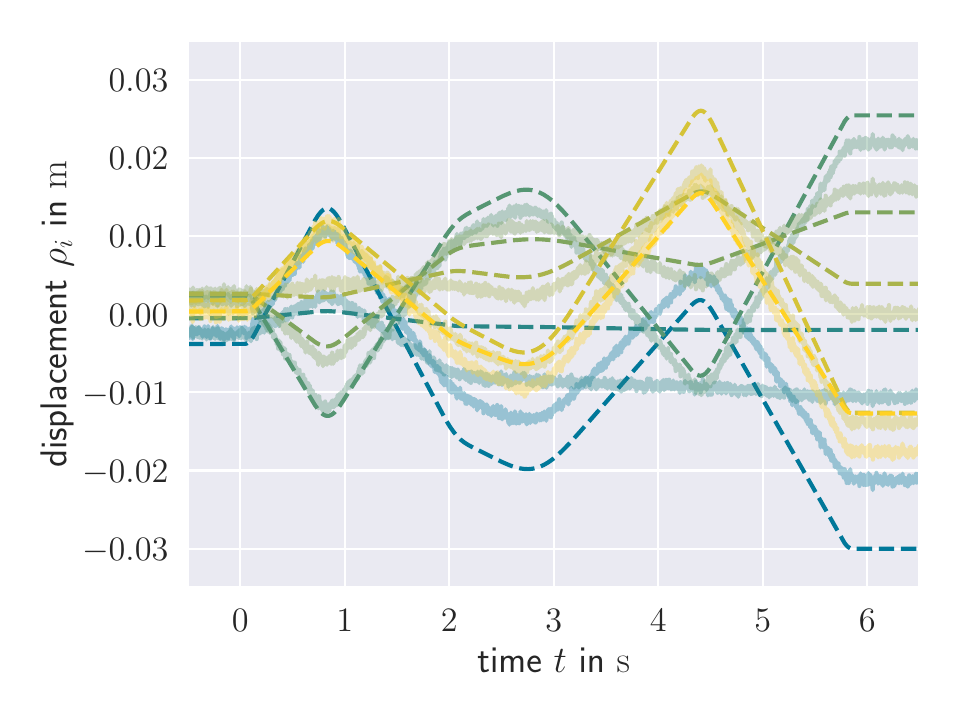}
        \put(42,78.25){\textit{robot\_D}}
    \end{overpic}
    \hfill
    \vspace*{-0.75em}
    \caption{
        Displacement-control without considering the non-constant distance $d_i$.
        The control outputs show that without using the appropriate encoder-decoder architecture, \textit{i.e.}, \eqref{eq:encoder-decoder} instead of \eqref{eq:encoder-decoder_general}, the desired path cannot be tracked accurately.
        Increasing the proportional gain or including an integrator might not be ideal, especially if this can be accomplished with \eqref{eq:encoder-decoder_general} as shown in Fig.~\ref{fig:results_closed_loop_advance}.
        }
    \label{fig:results_closed_loop_uncompensated}
    \vspace*{-1.75em}
\end{figure*}

Figure~\ref{fig:results_closed_loop_advance} shows the open-loop behavior without noise, open-loop behavior with added noise, and the closed-loop behavior for all five displacement-actuated continuum robots listed in Table~\ref{tab:design_parameters_evaluation}.
The controller output follows the desired transformed displacement of the respective target robot.
While the results in Fig.~\ref{fig:results_closed_loop_advance} utilize the more general definition of the encoder-decoder architecture \eqref{eq:encoder-decoder_general}, the results in Fig.~\ref{fig:results_closed_loop_uncompensated} are achieved using \eqref{eq:encoder-decoder}.

\section{Discussion and Future Work}

The alternative derivation highlights the importance of the displacement representation \eqref{eq:rho} over the relation to the standard Clarke transformation matrix $\boldsymbol{M}_\text{Clarke} \in \mathbb{R}^{3 \times 3}$ commonly used in the literature.
While no analogy to electrical engineering is used for the alternative derivation, possible synergy effects could be overlooked, and the relation between $\boldsymbol{M}_\text{Clarke}$ and $\MP$ is left in the dark.
Possible synergy effects are discussed in \cite{GrassmannBurgner-Kahrs_ICRA_EA_2024, GrassmannSenykBurgner-Kahrs_arXiv_2024}.

Generating joint values via rejection sampling has been shown in \cite{GrassmannSenykBurgner-Kahrs_arXiv_2024} to be inefficient even for the most simple symmetric case with $n = 3$.
To overcome the inefficiency, we combine two strategies; exploiting the geometric meaning of Clarke coordinates described in \cite{GrassmannSenykBurgner-Kahrs_arXiv_2024} and utilizing a simple surrogate displacement-actuated continuum robot.
Using an encoder-decoder architecture, feasible joint values for a target robot can be sampled.
This sampling method is branch-less, vectorizable, and rejection-free, \textit{i.e.}, all samples are \SI{100}{\%} correct. 

Transforming generated trajectories for a simple surrogate robot like \textit{robot\_0} using the encoder-decoder architecture is computationally cheaper than recomputing trajectories for the target robot.
The encoder-decoder architecture simplifies to a vector-matrix multiplication at each time step. 
In contrast, the trajectory generator \cite{GrassmannBurgner-Kahrs_RAL_2019} has five stages including the computation of octic polynomial functions, branching between three phases of the trapezoidal-like velocity profile, blending, and rescaling.
To improve trajectory generation, generating the desired trajectories directly on the two-dimensional manifold using the Clarke coordinates is desirable, where the kinematic constraints must be expressed in relation to the Clarke coordinates.

Furthermore, since superposition, \textit{i.e.}, scaling and addition, will not change the smoothness of the resulted trajectory, the $\mathcal{C}^{4}$-smooth trajectory for the surrogate robot remains a $\mathcal{C}^{4}$-smooth trajectory for the target robot.
However, different robot types have different requirements for the smoothness of the trajectory. 
For example, for rigid serial-kinematic robots, $\mathcal{C}^3$ smoothness is required \cite{BiagiottiMelchiorri_Book_2008}, whereas, for robots with flexible elements such as the Panda robot \cite{HaddadinHaddadin_et_al_RAM_2022}, a $\mathcal{C}^4$-smooth trajectory is mandatory \cite{DeLucaBook_HOR_2016}.
The requirements for continuum and soft robots are unknown yet and more work in this direction is needed.

The controller scheme illustrated in Fig.~\ref{fig:controller} utilizes the forward and inverse Clarke transforms.
For constant distances $d_i$ and constant length $l$, the controller gains are scaled resulting in different steady-state errors and noise sensitivity.
Constant factors in the Clarke transform can be taken into account in the controller gains.
That is why only one value for length $l$ in Table~\ref{tab:design_parameters_evaluation} is considered in the evaluation.
Furthermore, due to $\diag\left(1/d_i\right)$, the sensitivity to noise is higher for \textit{robot\_B}, \textit{robot\_C}, and \textit{robot\_D} as shown in Fig.~\ref{fig:results_closed_loop_advance}.
Tuning the controller gains will meditate this.
However, for uncompensated non-constant distance $d_i$, the steady-state error is significantly bigger and the PD controller scheme cannot accurately follow the trajectory as shown in Fig.~\ref{fig:results_closed_loop_uncompensated}, \textit{cf.}, Fig.~\ref{fig:results_closed_loop_advance}.
To account for non-constant distance $d_i$, we utilize \eqref{eq:encoder-decoder_general} instead of \eqref{eq:encoder-decoder}.
For future work, a model-based PD controller based on the kinematic design parameters is desirable to circumvent tuning of the controller gains.

The use of the proposed encoder-decoder architecture conveniently allows to reuse of a well-established robot morphology to sample feasible joint values, generate trajectories, and control the joint values.
It should be noted that the derivation of the encoder and decoder are model-based rather than learned using machine learning approaches.
Therefore, the transformation is geometrically exact and relies only on the kinematic design parameters as listed in Table~\ref{tab:design_parameters_evaluation} and used in \eqref{eq:encoder-decoder_general}.
Furthermore, note that, considering \eqref{eq:MP_inverse_pseudo}, forward \eqref{eq:robot_dependent_mapping_forward} and inverse robot-dependent mapping \eqref{eq:robot_dependent_mapping_inverse} are provided too.
Both representing the encoder and decoder part are useful in frameworks assuming constant curvature.

\subsection{Limitations}

It is assumed that the displacement-actuated joints can be used to pull and push.
However, for tendon-driven continuum robots, for instance, tendons can only be pulled.
Therefore, the location of the joints, \textit{i.e.}, location of the tendon holes, influences the motion capabilities of this type of continuum robot.
Further investigation on an appropriate clipping, shift, or storing approach is necessary to account for negative values related to pushing the tendons.

Regarding the encoder-decoder architecture, latent-space variables can be identified.
Using \eqref{eq:encoder-decoder}, the latent-space variables are the Clarke coordinates.
However, for the more general case using \eqref{eq:encoder-decoder_general}, the latent-space variables are not the Clarke coordinates.
To reconstruct some sort of Clarke coordinates, a function $f\!\left(d_i\right)$ considering all distances $d_i$ could be used.
For future work, the choice of the function $f\!\left(d_i\right)$ should be investigated to link the latent space variables back to a virtual displacement.
We kindly refer to \cite{GrassmannSenykBurgner-Kahrs_arXiv_2024, FirdausVadali_AIR_2023} on the concept of virtual displacement.
The choice of function $f\!\left(d_i\right)$ is important to clip the Clarke coordinates, \textit{e.g.}, to consider joint limits or saturation for PID controllers.

\subsection{Possible Applications}

\begin{figure}[thb]
    \centering
    \vspace*{0.5em}
    \includegraphics[width=0.35\columnwidth]{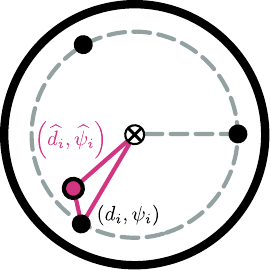}
    \caption{
        Uncertainty of the joint location results in asymmetric joint location.
        The polar coordinates $\left(d_i, \psi_i\right)$ refers to assumed location, whereas polar coordinates $\left(\widehat{d}_i, \widehat{\psi}_i\right)$ is the true location.
        }
    \vspace*{-1.5em}
    \label{fig:uncertaint-joint-location}
\end{figure}

Error propagation is a possible application.
Our approach can be used to analyze the propagation of error due to uncertainty in the joint location, see Fig.~\ref{fig:uncertaint-joint-location} for visual aid.
Furthermore, it would be possible to account for this error by accounting for the optimized value of the joint locations in the encoder-decoder architecture \eqref{eq:encoder-decoder_general}, within a piece-wise constant curvature framework.
However, an extension of Clarke transform to include twist to represent displacement-actuated continuum robot with \SI{3}{dof} per segment, \textit{e.g.,} \cite{GrassmannBurgner-Kahrs_et_al_Frontiers_2022, LeiSugaharaTakeda_ARK_2022}, would be desirable.

The consideration of asymmetric joint location opens new possibilities in designing physical hardware.
For example, for certain applications, it would be necessary to integrate working channels to deploy instruments for instance.
Those working channels cannot be used to realize a displacement-actuated joint.
In this case, an asymmetric joint location is desirable.
Another example is the use of additional displacement-actuated joints to increase the capability to withstand unwanted bending due to external forces or gravity.
Both examples are not mutually exclusive.
Figure~\ref{fig:non-symmetric_joint-location} illustrates possible designs.

The encoder-decoder architecture allows the resort to a surrogate robot with a well-established morphology.
As a use case, one might reinforce a well-established design with $n = 4$ with two more displacement-actuated joints to withstand gravity, see Fig.~\ref{fig:non-symmetric_joint-location}.
The encoder-decoder architecture is a key approach to reuse developed approaches for the well-established design.
For instance, the joint values generated by a planner can be transformed to joint values of the new design without reworking the developed planner, \textit{cf.} Fig.~\ref{fig:pipline}.

In fact, the encoder-decoder architecture can be used to translate frameworks for different robot morphologies.
For example, the model-based controller by Della Santina \textit{et al.} \cite{DellaSantinaBicchiRus_RAL_2020} is designed for a soft robot with four bellows.
The proposed encoder-decoder architecture can be used to consider a soft robot with five bellows instead of four bellows.
Obviously, certain parameters of the controller need to be tuned to the target robot.
Another example, by investing slightly more work in the adaptation, the controller by Della Santina \textit{et al.} \cite{DellaSantinaBicchiRus_RAL_2020} can be adapted to a tendon-driven continuum robot with five tendons.
This example illustrates that the Clarke transform can facilitate knowledge transfer between research fields.
A more straightforward example is the knowledge transfer between approaches for well-established designs with $n = 3$ and $n = 4$ symmetric arranged joints.
\section{Conclusions}

We utilize the Clarke transform and relax the assumption on the joint locations to an arbitrary number of joints and arbitrary joint locations.
We show how to generate feasible joint values for the target robot using a surrogate robot that can have a well-established robot morphology, \textit{e.g.}, three joints symmetrically arranged around the center-line.
One of the key approaches is the proposed encoder-decoder architecture to map the joint values of a surrogate robot to the target robot's joint values.
Our work allows the consideration of asymmetrically arranged joints opening the possibility to extend mechanical designs and the investigation of error propagation due to uncertainties.

\addcontentsline{toc}{section}{REFERENCES}
\bibliographystyle{IEEEtran}
\bibliography{IEEEabrv, references}

\begin{thebibliography}{10}
\providecommand{\url}[1]{#1}
\csname url@samestyle\endcsname
\providecommand{\newblock}{\relax}
\providecommand{\bibinfo}[2]{#2}
\providecommand{\BIBentrySTDinterwordspacing}{\spaceskip=0pt\relax}
\providecommand{\BIBentryALTinterwordstretchfactor}{4}
\providecommand{\BIBentryALTinterwordspacing}{\spaceskip=\fontdimen2\font plus
\BIBentryALTinterwordstretchfactor\fontdimen3\font minus \fontdimen4\font\relax}
\providecommand{\BIBforeignlanguage}[2]{{%
\expandafter\ifx\csname l@#1\endcsname\relax
\typeout{** WARNING: IEEEtran.bst: No hyphenation pattern has been}%
\typeout{** loaded for the language `#1'. Using the pattern for}%
\typeout{** the default language instead.}%
\else
\language=\csname l@#1\endcsname
\fi
#2}}
\providecommand{\BIBdecl}{\relax}
\BIBdecl

\bibitem{Burgner-KahrsRuckerChoset_TRO_2015}
J.~{Burgner-Kahrs}, D.~C. {Rucker}, and H.~{Choset}, ``\href{https://doi.org/10.1109/TRO.2015.2489500}{Continuum Robots for Medical Applications: A Survey},'' \emph{IEEE Transactions on Robotics}, vol.~31, no.~6, pp. 1261--1280, 2015.

\bibitem{DupontRucker_et_al_JPROC_2022}
P.~E. Dupont, N.~Simaan, H.~Choset, and D.~C. Rucker, ``\href{https://doi.org/10.1109/JPROC.2022.3141338}{Continuum robots for medical interventions},'' \emph{Proceedings of the IEEE}, vol. 110, no.~7, pp. 847--870, 2022.

\bibitem{DongKell_et_al_JMP_2019}
X.~Dong, D.~Palmer, D.~Axinte, and J.~Kell, ``\href{https://doi.org/10.1016/j.jmapro.2019.01.024}{In-situ repair/maintenance with a continuum robotic machine tool in confined space},'' \emph{Journal of Manufacturing Processes}, vol.~38, pp. 313--318, 2019.

\bibitem{RussoAxinte_et_al_AIS_2023}
M.~Russo, S.~M.~H. Sadati, X.~Dong, A.~Mohammad, I.~D. Walker, C.~Bergeles, K.~Xu, and D.~A. Axinte, ``\href{https://doi.org/10.1002/aisy.202200367}{Continuum robots: An overview},'' \emph{Advanced Intelligent Systems}, vol.~5, no.~5, p. 2200367, 2023.

\bibitem{LuWang_et_al_AR_2020}
J.~Lu, F.~Du, F.~Yang, T.~Zhang, Y.~Lei, and J.~Wang, ``\href{https://doi.org/10.1080/01691864.2020.1812427}{Kinematic modeling of a class of n-tendon continuum manipulators},'' \emph{Advanced Robotics}, vol.~34, no.~19, pp. 1254--1271, 2020.

\bibitem{AllenAlbert_et_al_RoboSoft_2020}
T.~F. Allen, L.~Rupert, T.~R. Duggan, G.~Hein, and K.~Albert, ``\href{https://doi.org/10.1109/RoboSoft48309.2020.9116015}{Closed-form non-singular constant-curvature continuum manipulator kinematics},'' in \emph{IEEE International Conference on Soft Robotics}, 2020, pp. 410--416.

\bibitem{DalvandNahavandiHowe_Access_2022}
M.~M. Dalvand, S.~Nahavandi, and R.~D. Howe, ``\href{https://doi.org/10.1109/ACCESS.2022.3180047}{General Forward Kinematics for Tendon-Driven Continuum Robots},'' \emph{IEEE Access}, vol.~10, pp. 60\,330--60\,340, 2022.

\bibitem{DellaSantinaBicchiRus_RAL_2020}
C.~Della~Santina, A.~Bicchi, and D.~Rus, ``\href{https://doi.org/10.1109/LRA.2020.2967269}{On an improved state parametrization for soft robots with piecewise constant curvature and its use in model based control},'' \emph{IEEE Robotics and Automation Letters}, vol.~5, no.~2, pp. 1001--1008, 2020.

\bibitem{GrassmannSenykBurgner-Kahrs_arXiv_2024}
R.~M. Grassmann, A.~Senyk, and J.~Burgner-Kahrs, ``\href{https://doi.org/10.48550/arXiv.2409.16501}{Clarke Transform -- A Fundamental Tool for Continuum Robotics},'' \emph{arXiv preprint arXiv:2409.16501}, 2024.

\bibitem{GrassmannBurgner-Kahrs_ICRA_EA_2024}
R.~M. Grassmann and J.~Burgner-Kahrs, ``\href{https://doi.org/10.48550/arXiv.2409.13826}{Clarke Transform and Clarke Coordinates -- A New Kid on the Block for State Representation of Continuum Robots},'' in \emph{40th Anniversary of the IEEE International Conference on Robotics and Automation}, 2024.

\bibitem{Janaszek_PIE_2016}
M.~Janaszek, ``\href{https://bibliotekanauki.pl/articles/159814.pdf}{Extended Clarke transformation for n-phase systems},'' \emph{Prace Instytutu Elektrotechniki}, no. 274, pp. 5--26, 2016.

\bibitem{Willems_TOE_1969}
J.~L. Willems, ``\href{https://doi.org/10.1109/TE.1969.4320448}{Generalized Clarke components for polyphase networks},'' \emph{IEEE Transactions on Education}, vol.~12, no.~1, pp. 69--71, 1969.

\bibitem{RockhillLipo_IEMDC_2015}
A.~Rockhill and T.~Lipo, ``\href{https://doi.org/10.1109/IEMDC.2015.7409032}{A generalized transformation methodology for polyphase electric machines and networks},'' in \emph{IEEE International Electric Machines \& Drives Conference}, 2015, pp. 27--34.

\bibitem{GrassmannSenykBurgner-Kahrs_ICRA_2024}
R.~M. Grassmann, A.~Senyk, and J.~Burgner-Kahrs, ``\href{https://doi.org/10.1109/ICRA57147.2024.10610322}{On the Disentanglement of Tube Inequalities in Concentric Tube Continuum Robots},'' in \emph{IEEE International Conference on Robotics and Automation}, 2024.

\bibitem{GrassmannBurgner-Kahrs_RAL_2019}
R.~M. Grassmann and J.~Burgner-Kahrs, ``\href{https://doi.org/10.1109/LRA.2019.2931133}{Quaternion-Based Smooth Trajectory Generator for Via Poses in SE(3) Considering Kinematic Limits in Cartesian Space},'' \emph{IEEE Robotics and Automation Letters}, vol.~4, no.~4, pp. 4192--4199, 2019.

\bibitem{BiagiottiMelchiorri_Book_2008}
L.~Biagiotti and C.~Melchiorri, \emph{Trajectory planning for automatic machines and robots}.\hskip 1em plus 0.5em minus 0.4em\relax Springer Science \& Business Media, 2008.

\bibitem{HaddadinHaddadin_et_al_RAM_2022}
S.~Haddadin, S.~Parusel, L.~Johannsmeier, S.~Golz, S.~Gabl, F.~Walch, M.~Sabaghian, C.~Jaehne, L.~Hausperger, and S.~Haddadin, ``\href{https://doi.org/10.1109/MRA.2021.3138382}{The Franka Emika Robot: A Reference Platform for Robotics Research and Education},'' \emph{IEEE Robotics \& Automation Magazine}, 2022.

\bibitem{DeLucaBook_HOR_2016}
A.~De~Luca and W.~J. Book, ``\href{https://doi.org/10.1007/978-3-319-32552-1_11}{Robots with flexible elements},'' in \emph{Springer Handbook of Robotics}.\hskip 1em plus 0.5em minus 0.4em\relax Springer, 2016, pp. 243--282.

\bibitem{FirdausVadali_AIR_2023}
M.~Firdaus and M.~Vadali, ``\href{https://doi.org/10.1145/3610419.3610491}{Virtual Tendon-Based Inverse Kinematics of Tendon-Driven Flexible Continuum Manipulators},'' in \emph{International Conference on Advances in Robotics}, 2023, pp. 1--5.

\bibitem{GrassmannBurgner-Kahrs_et_al_Frontiers_2022}
R.~M. Grassmann, P.~Rao, Q.~Peyron, and J.~Burgner-Kahrs, ``\href{https://doi.org/10.3389/frobt.2022.873446}{FAS -- A Fully Actuated Segment for Tendon-Driven Continuum Robots},'' \emph{Frontiers in Robotics and AI}, vol.~9, 2022.

\bibitem{LeiSugaharaTakeda_ARK_2022}
Y.~Lei, Y.~Sugahara, and Y.~Takeda, ``\href{https://doi.org/10.1007/978-3-031-08140-8_25}{Design and inverse kinematics of a novel tendon-driven continuum manipulator capable of twisting motion},'' in \emph{International Symposium on Advances in Robot Kinematics}.\hskip 1em plus 0.5em minus 0.4em\relax Springer, 2022, pp. 228--236.

\end{thebibliography}

\end{document}